\documentclass[acmtog, nonacm]{acmart}

\makeatletter
\let\@authorsaddresses\@empty
\makeatother

\AtBeginDocument{%
  \providecommand\BibTeX{{%
    \normalfont B\kern-0.5em{\scshape i\kern-0.25em b}\kern-0.8em\TeX}}}

\usepackage{subfiles}
\usepackage{graphicx}
\usepackage{array}
\usepackage{multirow}
\usepackage{MnSymbol}
\usepackage{algorithm}
\usepackage{algpseudocode}
\usepackage{hyperref}

\renewcommand{\text}[1]{\oldtext{\normalfont\slshape #1}}
\newlength\myindent
\begin{document}

\title{BlockFusion: Expandable 3D Scene Generation using Latent Tri-plane Extrapolation}

\author{Zhennan Wu}
\affiliation{%
 \institution{The University of Tokyo}
 \country{Japan}
 }

\author{Yang Li}
\affiliation{%
 \institution{Tencent XR Vision Labs}
 \country{China}}
\authornote{ Yang Li (liyang@mi.t.u-tokyo.ac.jp) is the corresponding author} 

\author{Han Yan}
 
\affiliation{%
\institution{Shanghai Jiao Tong University}
\country{China}}
\authornote{The contributions of Han, and Ruikai were made during their internships at Tencent XR Vision Labs.}

\author{Taizhang Shang}
 
\affiliation{%
\institution{Tencent XR Vision Labs}
\country{China}}

\author{Weixuan Sun}
 
\affiliation{%
\institution{Tencent XR Vision Labs XR}
\country{China}}

\author{Senbo Wang}
 
\affiliation{%
\institution{Tencent XR Vision Labs}
\country{China}}

\author{Ruikai Cui}
 
\affiliation{%
\institution{ANU}
\country{Australia}}

\author{Weizhe Liu}
 
\affiliation{%
\institution{Tencent XR Vision Labs}
\country{China}}

\author{Hiroyuki Sato}
 
\affiliation{%
\institution{The University of Tokyo}
\country{Japan}}

\author{Hongdong Li}
 
\affiliation{%
\institution{ANU}
\country{Australia}}

\author{Pan Ji}
 
\affiliation{%
\institution{Tencent XR Vision Labs}
\country{China}}

\begin{abstract}
We present BlockFusion, a diffusion-based model that generates 3D scenes as unit blocks and seamlessly incorporates new blocks to extend the scene.
BlockFusion is trained using datasets of 3D blocks that are randomly cropped from complete 3D scene meshes.
Through per-block fitting, all training blocks are converted into the hybrid neural fields: 
with a tri-plane containing the geometry features, followed by a Multi-layer Perceptron (MLP) for decoding the signed distance values.
A variational auto-encoder is employed to compress the tri-planes into the latent tri-plane space, on which the denoising diffusion process is performed.  
Diffusion applied to the latent representations allows for high-quality and diverse 3D scene generation.

To expand a scene during generation, one needs only to append empty blocks to overlap with the current scene and extrapolate existing latent tri-planes to populate new blocks.
The extrapolation is done by conditioning the generation process with the feature samples from the overlapping tri-planes during the denoising iterations.
Latent tri-plane extrapolation produces semantically and geometrically meaningful transitions that harmoniously blend with the existing scene.
A 2D layout conditioning mechanism is used to control the placement and arrangement of scene elements.
Experimental results indicate that BlockFusion is capable of generating diverse, geometrically consistent and unbounded large 3D scenes with unprecedented high-quality shapes in both indoor and outdoor scenarios.
%

\end{abstract}

%



\begin{teaserfigure}
\includegraphics[width=1.\textwidth]{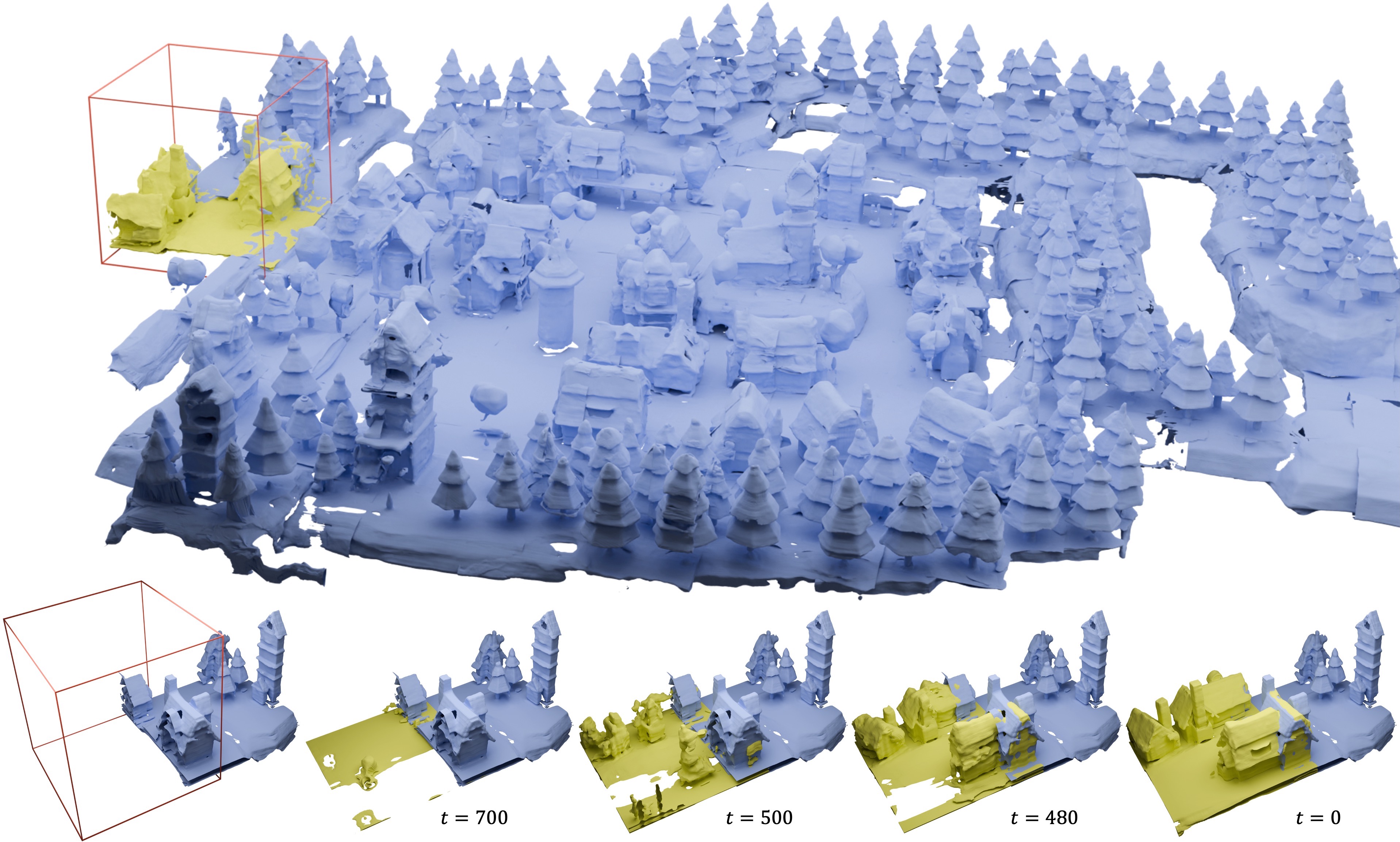}
\caption{
\textbf{BlockFusion generating a novel village.}
New block (red) is generated by extrapolating from existing ones.
Bottom row shows extrapolation steps.
}
\label{fig:teaser}
\end{teaserfigure}

\maketitle

\section{Introduction}

Generating large amount of high-quality 3D contents is key for many practical applications, including video-games, film-making, augmented and virtual reality (AR/VR).  
The increasing demand for high-quality digital contents has made 3D generation a significant topic of research. 
Recently, in the 2D domain, denoise diffusion models~\cite{sohl2015deep}, 
have demonstrated remarkable results in image synthesis and beyond, 
leading to the development of production-ready 2D generation tools, such as Stable Diffusion~\cite{stable_diffusion}, Midjourney, and Dall-E~\cite{ramesh2021zero}.
The success in 2D domain has significantly sparked interest in the development of 3D generation tools. 
A multitude of researches on 3D generation have been published recently, 
most notable works include DreamFusion~\cite{poole2022dreamfusion}, Rodin~\cite{rodin}, Get3D~\cite{gao2022get3d}, Zero123~\cite{zero123}, SyncDreamer~\cite{liu2023syncdreamer}, and LRM (\cite{hong2023lrm}), etc.
%

%
However, existing methods mainly focus on the generation of 3D content with fixed spatial extent (such as a small object of finite size).
In this paper, we investigate a relatively new yet increasingly important task: 
\textit{generating expandable (hence infinite) 3D scenes}.
This task is particularly valuable for video gaming industry, 
as it delivers an immersive gaming experience by allowing users to interact freely with the world without being restricted by a predetermined world boundary, as seen in open-world games.
Nonetheless, creating an unbounded and freely explorable scene is a non-trivial task.  Current practices typically rely on artists' manual labor, a time-consuming and costly process.
%

Generating expandable 3D scenes using diffusion models poses two major challenges:
First, 1) the generation of high-fidelity 3D shapes at the scene level is a difficult problem.
The variance in 3D scenes is orders of magnitude greater than in single objects.
A scene comprises basic objects, and the possibilities for arranging these objects are limitless.
This high level of diversity makes it difficult to approximate its distribution using diffusion probabilistic models.
Besides, 2) the expansion from an existing scene to a larger one is non-trival. 
The transition area between the old and new scenes needs to be both semantically and geometrically harmonious, adding another layer of complexity to the task.

Text2Room~\cite{hollein2023text2room} is the most closely related work to our task.  It employs a pre-trained 2D diffusion model to generate 2D images and lifts them to a 3D scene via the camera viewpoint and the estimated depth images. 
A scene is expanded by merging generated images from incrementally added camera viewpoints. Therefore, it is able to generate expandable 3D scenes with impressive texture results, though only at the room scale. 
However, since it critically relies on a monocular depth prediction, a poor depth prediction will lead to distorted geometry with missing details.
In addition, the way it expands a scene (i.e., by leveraging a moving perspective camera) makes it difficult to be extended beyond the room scale. 
This is because a perspective camera is prone to occlusion. For instance, when the camera passes through a wall, the continuity of the image can be disrupted by occlusion, which could also lead to discontinuities in the generated 3D shapes.

Instead of generating 3D through 2D image lifting,
another research direction involves directly learning to produce 3D data, 
using supervision from either 3D shape ground truths or posed multi-view images.
Notable methods include EG3D~\cite{EG3D}, Rodin~\cite{rodin}, and Get3D~\cite{gao2022get3d}, etc.
These approaches represent 3D data with a continuous hybrid neural field architecture, typically consisting of a tri-plane and an MLP decoder.
The tri-plane, originally introduced in EG3D~\cite{EG3D} and~\cite{peng2020convolutional}, is a tensor used to factorize the dense 3D volume grid. 
It is built on three axis-aligned 2D planes: the XY, YZ, and XZ planes.
The MLP decoder converts the tri-plane feature into a continuous value representing the scene, which could be occupancy, signed distance field (SDF), radiance field~\cite{nerf}, etc.
Tri-plane is significantly more compact and computationally efficient than a full 3D tensor and conducive to generative architectures developed for 2D image synthesis. 
This has been a key factor in making high-quality direct 3D data generation possible.\\
%

%
In this paper, we develop a tri-plane diffusion based approach to generate expandable 3D scenes.
Our method is called BlockFusion.
It generates 3D scenes in the form of cubic blocks and extends the scene in a straightforward sliding-block way.
To generate high-quality 3D shapes, we directly train BlockDiffusion on 3D scene datasets.
For network training, we randomly crop complete 3D scenes into incomplete 3D blocks with fixed sizes.
We run per-block fitting to convert all training blocks into tri-planes, which we call the raw tri-planes.
We found that directly training diffusion on raw tri-planes results in undesirable collapsed shapes. 
This issue is possibly caused by the high redundancy in the raw tri-planes and the substantial shape variance in the data.
Inspired by stable diffusion~\cite{stable_diffusion}, we apply an auto-encoder to compress the raw tri-planes into a latent tri-plane space to run diffusion.
The latent tri-plane space is significantly more compact and computationally efficient than the raw tri-plane while maintaining similar representation power.
In contrast to previous work, 
tri-plane diffusion on such a latent representation allows for the first time to reach high-quality and diverse 3D shape generation at scene level.
%

To expand a scene, we add empty blocks to overlap with the current scene and extrapolate existing tri-planes to populate the new blocks.
Specifically, the extrapolation is done by conditioning the generation process with the feature samples from the overlapping tri-planes during the reverse diffusion iterations.
The extrapolation is also carried out in the latent tri-plane space.
This process produces semantically and geometrically meaningful transitions that seamlessly blend with the existing scene, 
ensuring a coherent and visually pleasing scene expansion. 
%

To provide users with more control over the generation process, 
we introduce a 2D layout conditioning mechanism, 
which allows users to precisely determine the placement and arrangement of elements by manipulating 2D object bounding boxes.
We also demonstrate that the color and texture of the scenes can be created using an off-the-shelf texture generation tool, thereby increasing the visual allure of the scene.

To summarize, BlockFusion presents  
1) a generalizable, high-quality 3D generation model based on latent tri-plane diffusion,
2) a latent tri-plane extrapolation mechanism that allows harmonious scene expansion,
and 3) a 2D layout condition mechanism for precise control over scene generation.
Experimental results indicate that BlockFusion is capable of generating diverse, 
geometrically consistent and unbounded large 3D scenes with unprecedented high-quality shapes in both indoor and outdoor scenarios.

\section{Related Work}

\subsection{Diffusion models}
Starting from Gaussian noise samples, Diffusion probabilistic models \cite{sohl2015deep,ho2020denoising} generate clear images by learning to progressively remove noise from the original noise sample. 
Recent advances in diffusion models
\cite{nichol2021improved,dhariwal2021diffusion,ramesh2022hierarchical,saharia2022photorealistic} have demonstrated unprecedented capabilities in synthesizing high-quality and diverse images.
Nonetheless, training diffusion models directly in high-resolution pixel space can be computationally prohibitive.
Latent diffusion models (LDMs)~\cite{rombach2022high} address this
issue with a two-stage approach: they first compress the image through an auto-encoder and then apply diffusion models on smaller spatial representations in the latent space.
Diffusion models can be trained with guiding information
(e.g., text prompt, semantic layout, category label) to facilitate personalization, customization, or task-specific image generation.
There are basically two ways of manipulating generated content.
The first is realized through training a new model from scratch   or finetuning a pretrained diffusion model, adding various conditioning controls, e.g., sketch, depth, segmentation, 
\cite{wang2022pretraining,ramesh2022hierarchical,rombach2022high,rombach2022high,nichol2021glide,avrahami2023spatext,brooks2023instructpix2pix,wang2022semantic,zhang2023adding,li2023gligen,gal2022image,ruiz2023dreambooth,voynov2023sketch,bashkirova2023masksketch,huang2023composer,mou2023t2i}.
This approach requires extensive dataset building and extra computational consumption.
The other line of methods adapts pretrained model and
add some controlled generation capability during inference. With only slight modification to the generative process, \cite{tumanyan2023plug,hertz2022prompt,avrahami2022blended,bar2023multidiffusion} examine a wide variety of controlling diffusion models in a training/finetuning-free way.

\subsection{3D shape generation}
The success of 2D generation tools based on diffusion models, notably Stable Diffusion\cite{rombach2022high}, Midjourney, and Dall-E\cite{ramesh2021zero}, has significantly sparked interest in the development of 3D generation tools. There are two main streams for this task: the methods that lift 2D (generated) images into 3D models, and the methods that directly run diffusion on 3D data.
A thorough review of diffusion models for visual computing is available in~\cite{po2023diffusion_survey}.

\smallskip
\noindent
\textbf{2D-lifting methods.}
DreamFusion~\cite{poole2022dreamfusion} optimizes a Neural Radiance Field~\cite{nerf} using the Score Distillation Sampling (SDS) loss, 
which distills prior knowledge from 2D image diffusion models into the volume rendering output of the NeRF. 
Magic3D~\cite{lin2023magic3d} adopts an SDS loss-based second stage to further refine the mesh extracted from DreamFusion. 
SDS-based approaches demonstrate impressive results. 
However, they typically require hours of optimization and struggle with maintaining shape consistency, leading to a phenomenon called the Janus-face problem~\cite{poole2022dreamfusion}.
Several methods have been developed that focus on the direct generation of consistency-enhanced multi-view 2D images, and these techniques reconstruct 3D shapes from the generated multi-view images.
Zero123~\cite{zero123} fine-tunes Stable Diffusion model~\cite{stable_diffusion} to generate novel views by conditioning on the input image and camera transformation.
One2345~\cite{One-2-3-45} converts the multi-view image from Zero123 to 3D using an SDF-based neural surface reconstruction method.
One2345++~\cite{One-2-3-45++} fine-tunes a 2D diffusion model for consistent multi-view image generation, and then elevating these images to 3D with the aid of multi-view conditioned 3D diffusion models.
Syncdreamer~\cite{liu2023syncdreamer} and Consistnet~\cite{yang2023consistnet} synchronize multi-view image generation process by explicitly correlating features in 3D space.
Wonder3d~\cite{long2023wonder3d} improves the generation fidelity by introducing a cross-domain diffusion model that generates multi-view normal maps in addition to the color images.
LRM~\cite{hong2023lrm} treats the single-image-to-3D problem as a reconstruction problem and solves it using Transformer in a deterministic way. 
However, LRM can lead to blurry and washed-out textures for unseen parts of objects due to mode averaging.
To address this issue, Instant3D~\cite{li2023instant3d} inputs multi-view consistent images into LRM to infer geometry and textures for unseen parts.
DMV3D~\cite{xu2023dmv3d} employs LRM as a multi-view denoiser, which iteratively produces a cleaner tri-plane NeRF from noisy sparsely posed multi-view images.

\begin{figure*}[!t]
\includegraphics[width=\textwidth]{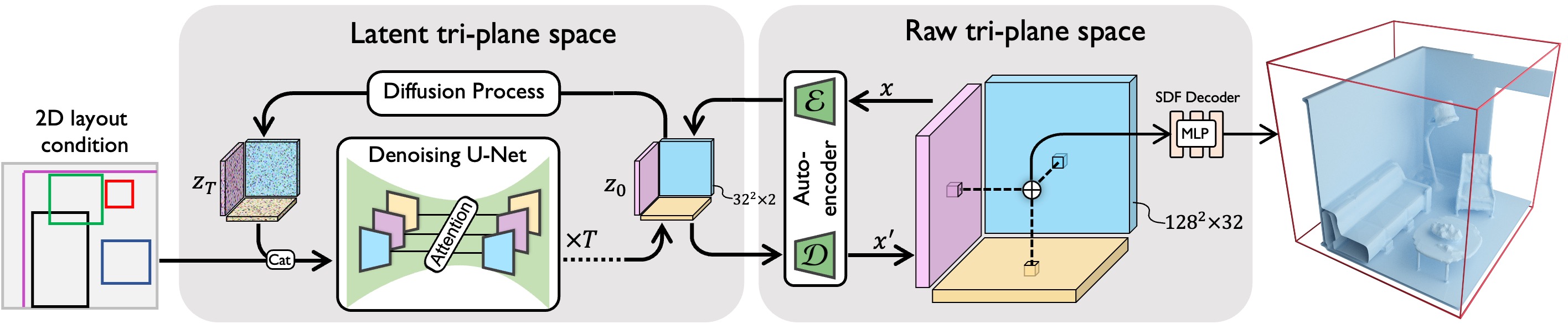} 
\caption{  
\textbf{BlockFusion training pipeline.}
The training contains three steps:
First, \textbf{1)} the training 3D blocks are converted to raw tri-planes via per-block shape fitting, c.f. Sec.~\ref{sec:trisdf}. 
Then, \textbf{2)} an auto-encoder compresses the raw tri-planes into a more compact latent tri-plane space, c.f. Sec.~\ref{sec:vae}.
Lastly, \textbf{3)} DDPM is trained to approximate the distributions of latent tri-planes, and during this process, layout control can also be integrated, c.f. Sec.~\ref{sec:diffusion}.
}
\label{fig:pipeline}
\end{figure*}

\smallskip
\noindent
\textbf{3D diffusion models.}
Another line of research involves directly training diffusion models to generate 3D shapes.
As the supervision comes directly from 3D shape ground truth or posed multi-view images, 
the generated results typically exhibit superior geometric quality compared to those from 2D diffusion-based methods.
These methods can be categorized based on the type of 3D representations they employ, including:
polygon meshes~\cite{gao2022get3d,Liu2023MeshDiffusion}, 
point clouds~\cite{nichol2022point-e,zeng2022lion}, 
explicit 3D grids holding occupancy or SDF values~\cite{lasdiffusion,One-2-3-45++}, 
or neural fields~\cite{rodin,nfd,SSDNerf,muller2023diffrf,jun2023shap-e, dmv3d, erkocc2023hyperdiffusion,diffusion-sdf}.
The hybrid neural field, which incorporates a tri-plane followed by a neural decoder, has been widely adopted in 3D diffusion models due to its computational efficiency.
Rodin~\cite{rodin} first fits tri-plane NeRFs for a human upper body dataset, and then uses a two-stage coarse-to-fine diffusion model to generate the corresponding tri-planes.
Similarly, NFD~\cite{nfd} trains a tri-plane diffusion model for 3D data parametrized via occupancy values.
SSDNerf~\cite{SSDNerf} merge tri-plane fitting and generation into a single-stage pipeline. 
However, in practice, tri-plane diffusion is still challenging to train due to its high dimensionality and irregularity. 
Existing methods only demonstrated simple cases with small data varieties, i.e., Rodin for canonicalized human upper-body dataset~\cite{fake-it-till-make-it}, and NFD and SSDNeRF for single-category objects in ShapeNet~\cite{chang2015shapenet}. 
This paper follows this line of research but introduces a major change: we use an auto-encoder to compress the tri-plane into a highly compact latent tri-plane space for diffusion. 
We demonstrate that this approach significantly improves the stability, generalizability, and output quality of tri-plane diffusion.

\subsection{3D scene generation}
Generating 3D scenes presents a more substantial challenge than generating single objects.
This is because scenes are geometrically more complex than individual objects, and they cannot be contained in a fixed spatial size.
Object retrieval-based approaches assume there is a database of objects,
and they arrange the retrieved objects to fill an empty scene, as seen in Diffuscene~\cite{tang2023diffuscene} and Sceneformer~\cite{wang2021sceneformer},
consequently, the synthesized scene can not contain novel elements that do not exist in the database.
Text2Room~\cite{hollein2023text2room} is the first method that uses 2D diffusion model to build a 3D generation tool.
It first generates color and depth frames using 2D diffusion models, 
and then shift camera positions to generate new frames, which are integrated into a global map.
A similar approach for indoor scenarios can be found in SceneScape~\cite{SceneScape}.
To allow precise control over the contents generated in a scene,
ControlRoom3D\cite{schult23controlroom3d} and CTRL-ROOM\cite{fang2023ctrl} develop panorama-based room generation models that take 3D room layouts as input conditions.
CC3D~\cite{cc3d} utilizes a 3D layout to guide the SDS~\cite{poole2022dreamfusion} process.
Citygen~\cite{deng2023citygen} represents city scenes using the height map proxy, leading to a 2.5D scene generation.
PERF\cite{perf2023} proposes a novel panoramic view synthesis framework, thereby lifting 2D panorama up to 3D scene.
Other approaches, such as~\cite{Tang2023mvdiffusion,chen2023scenedreamer} focus on generating scenes with high-quality visual appearances.
Given a room scene mesh, MVDiffusion~\cite{Tang2023mvdiffusion} generates coherent multiview perspective images, which can be lifted to the 3D as the UV texture of the mesh.
SceneDreamer~\cite{chen2023scenedreamer} leverages in-the-wild 2D images to construct large scenes with photo-realistic volume rendering effects.
However, it still depends on 3D shapes represented by semantic height maps as input, consequently, the dimensions of the generated scenes are bounded by the dimensions of the input shapes.

In this paper, we address the fundamental challenge of generating an unbounded scene by developing an auto-regressive scene expansion algorithm based on tri-plane diffusion.

\section{Method}
BlockFusion generates scenes as blocks and expands scenes using a sliding-window progressive generation approach. 
Fig.~\ref{fig:pipeline} presents the training pipeline.
This section is organized as follows:
\begin{itemize}
   \item Sec.~\ref{sec:manifoldness} demonstrates how the training blocks are generated.
   \item Sec.~\ref{sec:trisdf}: we run per-block fitting to convert all training blocks into tri-planes, which we call the raw tri-planes.
   \item Sec.~\ref{sec:vae}: the raw tri-planes are compressed into a latent tri-plane space for efficient 3D representation.
   \item Sec:~\ref{sec:diffusion}: we train the diffusion model on the latent tri-plane space. 
   \item Sec.~\ref{sec:tri-expolation}: we leverage the pre-trained latent tri-plane diffusion model to expand a scene.
   \item Sec.~\ref{sec:pp}: a post-processing technique is adapted to reduce seams.
   \item Sec.~\ref{sec:large_scene}: large scenes are built by running BlockFusion progressively.
\end{itemize}

\subsection{Crop training scenes into 3D blocks} \label{sec:manifoldness}
\begin{figure}[!t]
\centering
\includegraphics[width=0.48\textwidth]{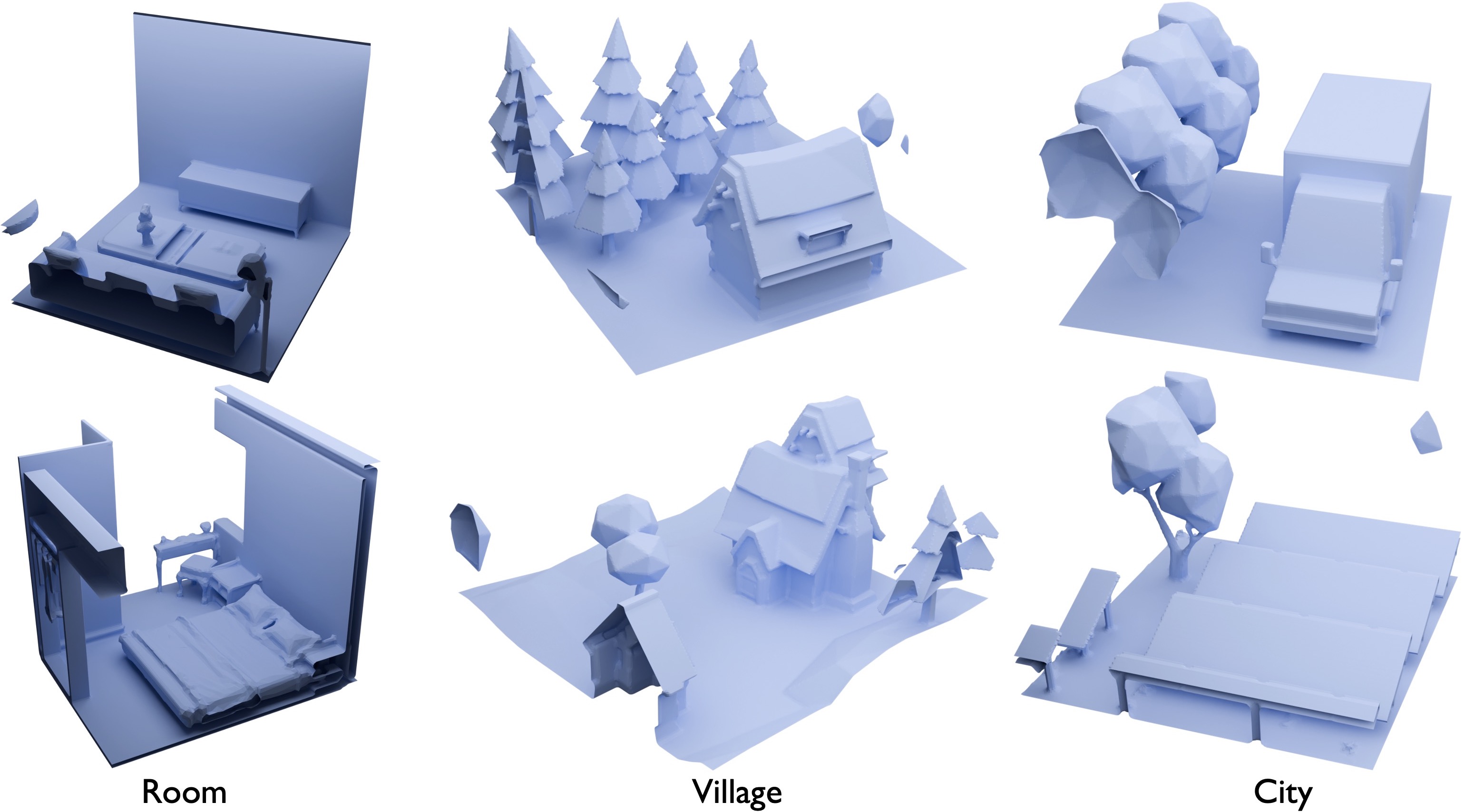}
\caption{Examples of randomly cropped 3D blocks.}
\label{fig:scene_crop}
\end{figure}

We use scene meshes for network training.
We convert scene meshes to water-tight meshes and then randomly crop the meshes into cubic blocks.
The size of the block is adjusted such that it is large enough to enclose major objects in the scene, e.g. beds in room scenes, or houses in outdoor scenes. 
Given that the blocks are randomly positioned within the scene, objects may be split by these blocks. 
In addition, the possible arrangements of objects within a block are limitless.
Considerably, the variance in such a randomly cropped shape dataset is much larger than that of a single object-centered dataset.
As a result, training diffusion on this type of data presents a greater challenge.
We test on three different types of scenes including room, city, and village.
Examples of training blocks can be found in Fig.~\ref{fig:scene_crop}.
In addition to the shapes, we also create a 2D layout map for each scene. 
The layout map is the ground plane projection of the objects, grouped by their categories. 
These layout maps can be used as input conditions for diffusion, 
so we also crop them accordingly. Examples can be seen in Fig.~\ref{fig:pipeline}.

\subsection{Raw tri-plane fitting}\label{sec:trisdf}

\medskip
\noindent
\textbf{Hybrid Neural SDF.}
We use the signed distance field (SDF) to represent the shape.
An SDF is a continuous distance function with values indicating the distance to the surface and signs indicating whether a point is inside or outside the object.
The final surface can be extracted via marching cubes.
The shape is reconstructed using the hybrid neural field structure, which consists of a tri-plane to hold the geometry feature and a  multiple layer perceptron (MLP) with parameter $\theta$ to decode the signed distance value.
The tri-plane is a tensor used to factorize the dense 3D volume grid.
It is built on three axis-aligned 2D planes: the XY, YZ, and XZ planes.
Formally, it reads $ x = \bigl\{x(i)|x(i)\in \mathbb{R}^{N^2\times C}, {i\in\{1,2,3\}} \bigr\}$, 
where $N^2$ is the plane resolution and $C$ is the dimension of the feature. 
Given a query point $p\in\mathbb{R}^3$, the function  $\Phi: \mathbb{R}^3 \mapsto \mathbb{R}$  outputs the signed distance value:
\begin{equation}
    \Phi(p) = \textnormal{MLP}_{\theta} \biggl(\bigoplus  _{i\in\{1,2,3\} } \textnormal{Interp}_{x(i)} \Bigl(\textnormal{Proj}_{x(i)} \bigl( p \bigr) \Bigr) \biggl)
\end{equation}
where $\textnormal{Proj}(\cdot)$ represents orthogonal point-to-plane projection,
$\textnormal{Interp}(\cdot)$ refers to bi-linear interpolation that queries feature vectors from each plane respectively, 
and $\bigoplus$ denotes element-wise addition.
The addition operation is performed along the feature dimension,  reducing the three feature vectors into a single final feature.

\medskip
\noindent
\textbf{Training points sampling.} \label{sec:point_sample}
Given the mesh of a training block, we sample on-surface points and off-surface points, and then compute the ground truth SDF values.
On-surface point set, denoted as  $\Omega_0$, is randomly sampled on the surface.
Their SDF values are equal to zero and we also compute the surface normal GT for each of these points.
Off-surface point set denoted as $\Omega$ is sampled uniformly at random inside the block.
To avoid incorrect distance values resulting from mesh cropping, 
the ground truth SDF values of the off-surface points are computed with respect to the original water-tight mesh.
We empirically found that the point set sizes $|\Omega|=100000$ and $|\Omega_0|=500000$ 
achieve solid shape-fitting results while maintaining optimization costs at a manageable level.
The XYZ coordinates of all sampled points are normalized to the range [-1, 1].

\medskip
\noindent
\textbf{Triplane Fitting.}
Our goal is to transform all training blocks into tri-planes, which will then be used for training our generative model.
Inspired by pioneering works on shape representation with neural field-based SDFs~\cite{SAL,deepsdf,eilknal_loss}, 
we jointly optimize tri-plane $x$ and MLP weights $\theta$ with the following geometry loss:
\begin{equation}\label{eqn:l-geo}
\mathcal{L}_{geo}=\mathcal{L}_{SDF}+\mathcal{L}_{Normal}+\mathcal{L}_{Eikonal}    
\end{equation}
The three terms are:
\begin{equation}\label{eqn:l-sdf}
    \mathcal{L}_{SDF} =\lambda_{1} \sum_{p\in \Omega_0} || \Phi(p) || + 
    \lambda_{2}\sum_{p\in \Omega}|| \Phi(p) - d_p ||
\end{equation}
\begin{equation}\label{eqn:l-normal}
    \mathcal{L}_{Normal} = \lambda_3  \sum_{p\in \Omega_0}|| 
 \nabla_p\Phi(p) -  \mathbf{\mathrm{n}}_p ||
\end{equation}
\begin{equation}\label{eqn:l-eikonal}
    \mathcal{L}_{Eikonal} =\lambda_4 \sum_{p\in \Omega_0}|| 
 |\nabla_p\Phi(p)| -  1 ||
\end{equation}
where $d_p$ and $\mathbf{\mathrm{n}}_p$ are ground truth SDF values and surface normal vector.
The gradient $\nabla_p\Phi(p) = [\frac{ \partial{\Phi(p)}}{ \partial{X}},\frac{ \partial{\Phi(p)}}{ \partial{Y}},\frac{ \partial{\Phi(p)}}{ \partial{Z}}]$ represents the direction of the steepest change in SDF. It can be computed using finite difference, e.g. the  partial derivative for the X-axis component reads
\begin{equation}
  \frac{ \partial{\Phi(p)}}{ \partial{X}} = \frac{\Phi(p + [\delta,0,0]) - \Phi(p - [\delta,0,0]) } {2\delta}
\end{equation}
 where $\delta$ is the step size.
The Eikonal loss constrains $|\nabla_p\Phi(p)|$ to be 1 almost everywhere, thus maintaining the intrinsic physical property of the signed distance function.
We adopt the MLP initialization trick as introduced in SAL~\cite{SAL}, 
which constrains the initial SDF output to roughly approximate a sphere.
This spherical geometry initialization technique significantly facilitates global convergence.
Empirically, the loss weights are set to $\lambda_{1}=100.0$, $\lambda_{2}=3.0$, $\lambda_{3}=1.0$, and $\lambda_{4}=0.5$ across all datasets.
The MLP is jointly trained with the tri-planes using a training subset consisting of 500 blocks. 
Upon convergence, the MLP is regarded as a generalizable SDF decoder.
Then we freeze MLP and optimize the tri-planes for all blocks in the training data. 
In this work, the output tri-plane size is set to $N^2=128^2, C=32$.
Following~\cite{yan2024frankenstein}, a tri-plane is optimized in a coarse-to-fine manner, i.e., the resolution is initialized with $8^2$ and gradually up-scaled to $128^2$. 
Compared to directly optimizing at the final resolution, this trick significantly improves fitting robustness and reduces running time.

\medskip
Now, we can convert a dataset of 3D blocks into a dataset of tri-planes with size $3\times128^2\times32$.
These tri-planes can faithfully reconstruct the 3D blocks, c.f. Fig.~\ref{fig:block_recon}.
We call them as the \textit{raw tri-planes}.

\subsection{Compressing to latent tri-plane space} \label{sec:vae}

\begin{figure}[!t]
\centering
\includegraphics[width=0.48\textwidth]{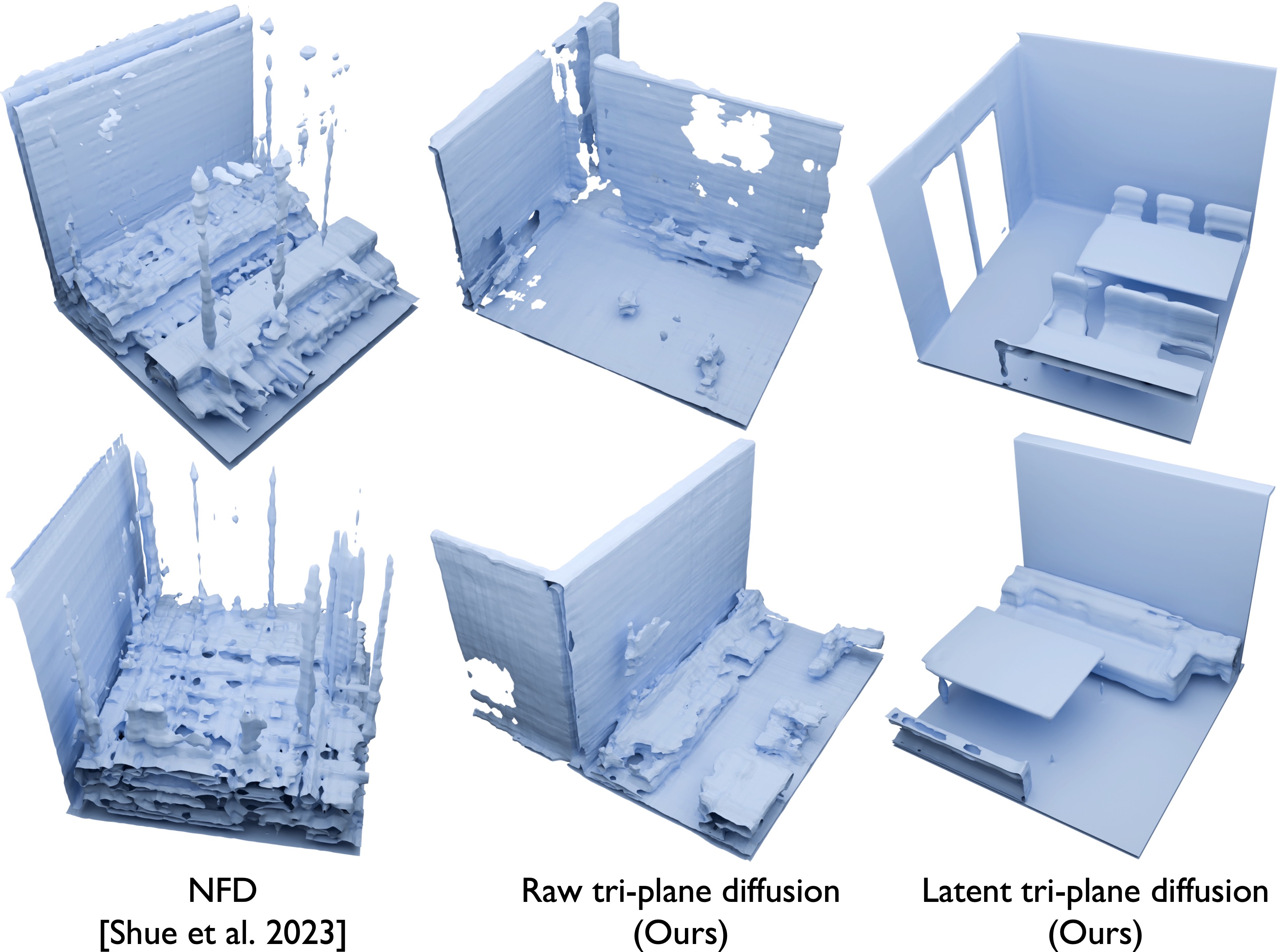}
\caption{
\textbf{Qualitative unconditioned block generation results.}
NFD~\cite{nfd} is also based on tri-plane diffusion. 
They utilize occupancy value to represent shapes, 
whereas ours employ SDF.
All three methods are trained on room blocks.
}
\label{fig:uncond_gen}
\end{figure}

Although our raw tri-planes can reconstruct high-quality shapes, 
we found that generating such tri-planes is significantly difficult.
Directly training diffusion models on such tri-planes leads to collapsed results, as shown in Fig.~\ref{fig:uncond_gen}.
We argue that there are mainly two reasons for this:
1) the raw tri-plane is highly redundant,
and 2) the shape diversity in our scene block dataset is too large. 
Although previous works like Rodin~\cite{rodin} and NFD~\cite{nfd} have proven the feasibility of diffusion on raw tri-planes, 
they only work on datasets with much smaller varieties,
i.e., Rodin for canonicalized human upper bodies,
and NFD for single-category objects from ShapeNet~\cite{chang2015shapenet}. 
When we attempted to retrain NFD on our scene blocks, it also failed to produce meaningful shapes, as shown in Fig.~\ref{fig:uncond_gen}.

We need to find a feature representation for 3D shapes that is compact, easy to train diffusion models, memory and computationally efficient, and capable of generalizing to large shape variations.
In the 2D scenario, Stable Diffusion~\cite{stable_diffusion} compresses raw images into a latent 2D feature space for diffusion. 
This approach results in a more robust model that generates higher-quality images. 
Inspired by Stable Diffusion, we train an auto-encoder to compress raw tri-planes into a latent tri-plane space with reduced resolution and feature channels.
Precisely, given a raw tri-plane $x\in{\mathbb{R}^{3\times{N^2}\times{C}}}$, 
the encoder $\mathcal{E}$ encodes $x$ into a latent representation $z=\mathcal{E}(x)$, 
and the decoder $\mathcal{D}$ reconstructs the raw tri-plane from the latent, 
giving $\hat{x}= \mathcal{D}(z) = \mathcal{D}(\mathcal{E}(x))$.

\medskip
\noindent
\textbf{Training objective} of the auto-encoder is shown as follows:
\begin{equation}
    \mathcal{L}_{AE} = \mathcal{L}_{rec}(x,\mathcal{D}(\mathcal{E}(x))) + \mathcal{L}_{KL}(x,\mathcal{D},\mathcal{E}) + \mathcal{L}_{geo} 
\end{equation}
where $\mathcal{L}_{rec}$ is a light $L_1$ norm applied between $x$ and its reconstruction $\mathcal{D}(\mathcal{E}(x))$.
$\mathcal{L}_{KL}$ is a the Kullback-Leibler-term between $q_{\mathcal{E}}(z|x) = \mathcal{N}(z; \mathcal{E}_{u}, \mathcal{E}_{\sigma^2} )$ and a standard normal distribution $ \mathcal{N}(\textbf{0},\textbf{I})$ as in a standard VAE~\cite{vae}.
To obtain high-fidelity shape reconstructions we only use a very small weight for $\mathcal{L}_{KL}$.
$\mathcal{L}_{geo}$ is the geometry loss defined in Eqn.~\ref{eqn:l-geo}.
It is assessed based on the same set of points as in Sec.~\ref{sec:point_sample}. 
Since the purpose is to learn a latent tri-plane that can faithfully represent the shape,
we rely on $L_{geo}$ as the dominate loss for training the auto-encoder.
Detailed structures of VAE are presented in appendix.  \\

The latent $z$ mantains a tri-plane structure with $ z = \bigl\{z(i)|z(i)\in \mathbb{R}^{n^2\times c}, {i\in\{1,2,3\}} \bigr\}$.
We call it as the \textit{latent tri-plane}.
This is in contrast to the previous work DiffusionSDF~\cite{diffusion-sdf},
which relies on an arbitrary one-dimensional latent vector $z$ to model its distribution autoregressively and thereby ignores much of the inherent 3D structure of $z$. 
Hence, our compression model preserves details of $x$ better (see Fig.~\ref{fig:block_recon}).
Empirically, the latent resolution is set to $n^2=32^2$.
And we investigate two different latent feature dimensions with $c=2$ and $c=16$.

\subsection{Latent Triplane Diffusion} \label{sec:diffusion}
With our trained tri-plane auto-encoder, comprising $\mathcal{E}$ and $\mathcal{D}$, 
we now have access to an efficient, low-dimensional latent tri-plane space where high-frequency, imperceptible details are abstracted away. 
In comparison to the raw tri-plane space, 
the latent tri-plane space is more suitable for likelihood-based generative models,
as they can now concentrate on the essential, semantic aspects of the data 
and train in a lower-dimensional, computationally much more efficient space. \\

\noindent
\medskip
\textbf{Background on Diffusion Probabilistic Models.} Diffusion Models are probabilistic models designed to learn a data distribution $z_0 \sim q(z_0)$ by gradually denoising a normally distributed variable. 
This process corresponds to learning the reverse operation of a fixed Markov Chain with a length of $T$. 
The inference process works by sampling a random noise $z_T$ and gradually denoising it until it reaches a meaningful latent $z_0$.
DDPM~\cite{ho2020denoising} defines a diffusion process that transform latent $z_0$ to white Gaussian noise $z_T \sim  \mathcal{N}(\textbf{0},\textbf{I})$ in $T$ time steps. Each step in the forward direction is given by:
\begin{equation}
q(z_1,...,z_T|z_{0}) = \prod_{t=1}^{T}q(z_t|z_{t-1})
\end{equation}
\begin{equation}\label{eqn:sample}
q(z_t|z_{t-1}) =  \mathcal{N}(z_t;\sqrt{1-\beta_t}z_{t-1},\beta_t\textbf{I})
\end{equation}
The noisy latent $z_t$ is obtained
by scaling the previous noise sample $z_{t-1}$ with $\sqrt{1-\beta_t}$ and adding Gaussian noise with variance $\beta_t$ at timestep $t$ .
During training, DDPM reverses the diffusion process, which is modeled by a neural network  $\Psi$ that predicts the parameters $\mu_{\Psi}(z_t,t)$ and $\Sigma_{\Psi}(z_t,t)$ of a Gaussian distribution.
\begin{equation}
p_\Psi(z_{t-1}|z_t) =  \mathcal{N}(z_{t-1};\mu_\Psi(z_t,t),\Sigma_\Psi(z_t,t))
\end{equation}
With $\alpha_t:=1-\beta_t$ and $\bar{\alpha}:=\prod_{s=0}^{t}a_s$, we can write marginal distribution:
\begin{equation}
q(z_t|z_0) = \mathcal{N}(z_t;\sqrt{\bar{\alpha}_t}z_0,(1-\bar{\alpha}_t)\textbf{I})
\end{equation}
\begin{equation}\label{eqn:addnoise}
z_t = \sqrt{\bar{\alpha}_t}z_0+\sqrt{1-\bar{\alpha}_t}\epsilon
\end{equation}
where $\epsilon \sim \mathcal{N}(\textbf{0},\textbf{I}) $.
Using Bayes theorem, one can calculate the posterior $q(z_{t-1}|z_t,z_0)$ in terms of $\tilde{\beta_t}$ and  $\tilde{\mu}_t(z_t,z_0)$ which are defined as follows:
\begin{equation}
\tilde{\beta_t}:=\frac{1-\bar{\alpha}_{t-1}}{1-\bar{\alpha}_{t}}\beta_t
\end{equation}
\begin{equation}\label{eqn:mu}
\tilde{\mu}_t(z_t,z_0) := \frac{\sqrt{\bar{\alpha}_{t-1}}\beta_t}{1-\bar{\alpha}_t}z_0 + \frac{\sqrt{\alpha_t}(1-\bar{\alpha}_{t-1})}{1-\bar{\alpha}_t}z_t
\end{equation}
\begin{equation}
q(z_{t-1}|z_t,z_0) = \mathcal{N}(z_{t-1};\tilde{\mu}_t(z_t,z_0),\tilde{\beta}_t\textbf{I})
\end{equation}
\medskip

There are different ways to parameterize $\mu_{\Psi}(z_t,t)$ in the prior. 
In this paper, we predict $z_0$ directly with a neural network $\Psi$. The prediction could be used in Eqn. \ref{eqn:mu} to produce $\mu_{\Psi}(z_t,t)$.
Specifically, with a uniformly sampled time step $t$ from $\{1, . . . , T\}$, we sample noise to obtain $z_t$ from input latent vector $z_0$.  
A time-conditioned denoising auto-encoder $\Psi$ learns to reconstruct $z_0$ from $z_t$.
The objective of latent tri-plane diffusion reads
\begin{equation}
\mathcal{L}_{LTD} = ||\Psi(z_t,\gamma(t))-z_0||_2
\end{equation}
where $\gamma(\cdot)$ is a positional encoding function and $||\cdot||_2$ is MSE loss. 
Since the forward process is fixed, $z_t$ can be efficiently obtained from $\mathcal{E}$ during training.
During test time, we iteratively denoise $z_T$ until we obtain the final output $z'$.  
$z'$ can be decoded to the raw tri-plane $x'$ with a single pass through $\mathcal{D}$.
Finally, the pretrained MLP decodes $x'$ to a dense SDF volume for shape extraction through marching cube.

\begin{figure}[!t]
\centering
\includegraphics[width=0.48\textwidth]{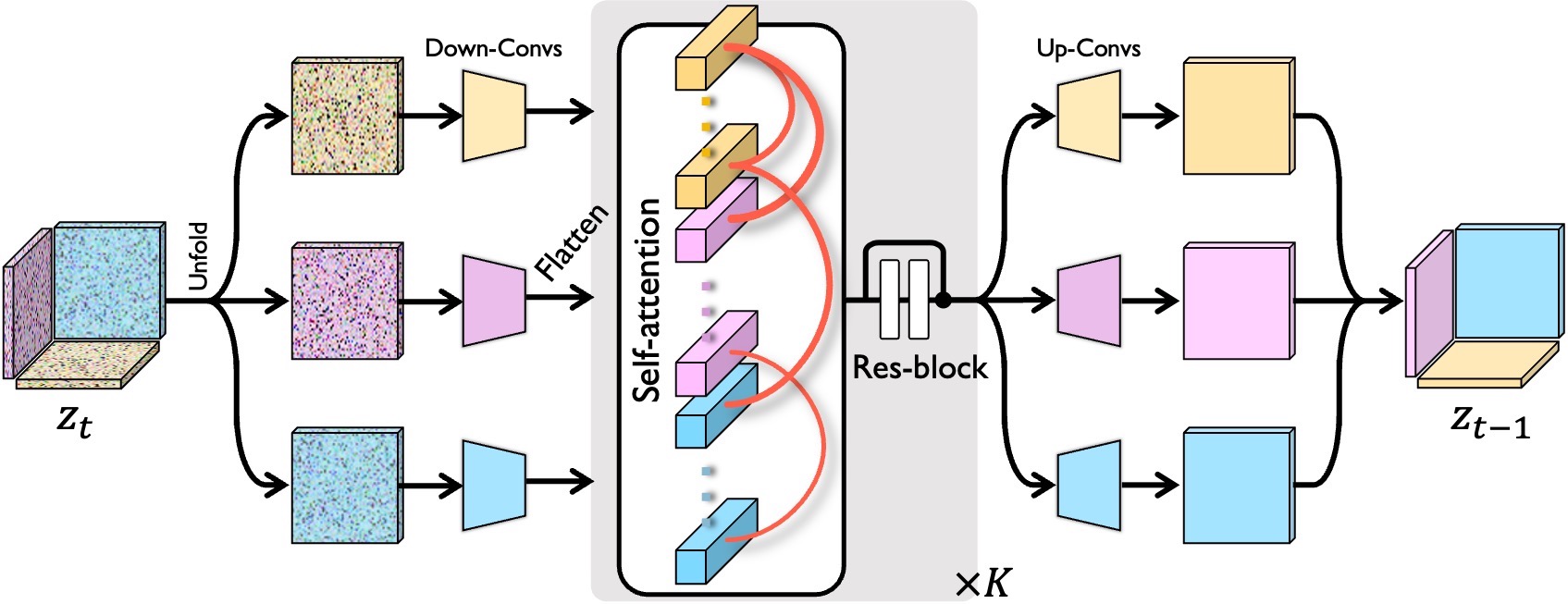}
\caption{ \textbf{3D aware denoising U-Net.}
The latent tri-plane is unfolded into three independent planes to run down-sampling convolutions.
After the down-sampling layers, the three feature maps are flattened into 1D tokens and concatenated together to forward through a sequence of self-attention~\cite{vaswani2017attention_is_all_need} and residual block by $K=6$ times.
Finally, the 1D array is reshaped into planes for up-sampling convolution and reassembled into the tri-plane structure.
}
\label{fig:3d-aware-backbone}
\end{figure}

\medskip
\noindent
\textbf{2D layout as user control.}
To control the generation process,
we add floor layout control by informing the model with 2D bounding box projections of objects.
The floor layout is converted into a feature map $l \in \mathbb{R}^{n^2 \times m}$, 
where $n^2$ represents the feature resolution (identical to the plane resolution in latent $z$), 
and the channel number $m$ corresponds to the total number of object categories.
Each channel consists of a binary image indicating whether or not an object class is placed. 
The loss of layout-conditioned latent tri-plane diffusion reads
\begin{equation}
\mathcal{L}_{c-LTD} = ||\Psi(z_t,\gamma(t),l)-z_0||_2
\end{equation}
In practice, $l$ is directly concatenated to three planes of $z_t$. 
In our experiments, we show that this type of conditioning successfully controls the arrangement of scene elements while still preserving variance in the generated shapes.

\medskip
\noindent
\textbf{3D aware denoising U-Net.} \label{sec:3d-aware}
The neural backbone $\Psi(\cdot)$ of our model is realized as a time-conditional U-Net.
The advantage of tri-planes is that we can treat them as 2D tensors and therefore apply efficient 2D convolutions.
However, naively running convolution on tri-planes does not produce satisfactory results, as the 3D relationships among the plane features are ignored.
$\Psi(\cdot)$ needs to incorporate operations that can account for the cross-plane feature relationships.
To address this, Rodin~\cite{rodin} introduces a 3D-aware convolution, which employs max-pooling and concatenation to associate features between planes based on their 3D correlations.
However, the simple max-pooling max-pooling selects the largest one and discards the rest, inevitably causing information loss.
In this work, we build $\Psi(\cdot)$ by leveraging the more powerful transformer to achieve cross-plane communication. 
The overall architecture of $\Psi(\cdot)$ is shown in Fig.~\ref{fig:3d-aware-backbone}.
This architecture enables effective 3D-aware feature learning.

\begin{figure}[!b]
\includegraphics[width=0.46\textwidth]{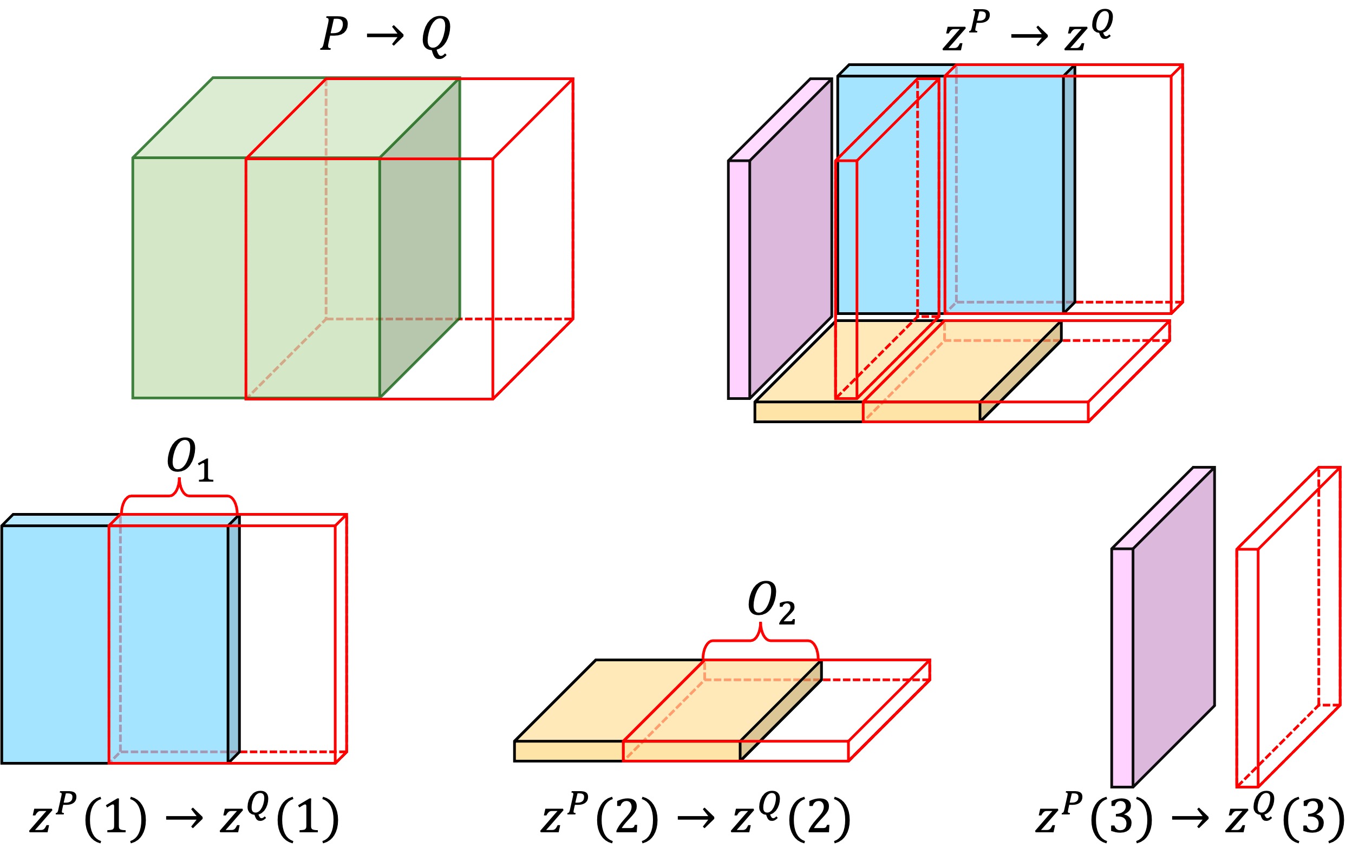} 
\caption{ \textbf{ Latent triplane extrapolation.}
Given the known block $P$  and the unknown block $Q$ ,
the goal is to extrapolate the known latent tri-plane $z^P$ to obtain the unknown tri-plane $z^Q$ (top row).
This tri-plane extrapolation is factored into the extrapolation of three 2D planes separately (bottom row).
}
\label{fig:pipe_triplane_extrapolation}
\end{figure}

\subsection{Latent tri-plane Extrapolation} \label{sec:tri-expolation}
Repaint~\cite{lugmayr2022repaint} demonstrate impressive image inpainting and extrapolation results using a pre-trained diffusion model. 
Their key idea is to synchronize the denoising process of the unknown pixels using the noised version of the known pixels.
Inspired by Repaint, we leverage our pre-trained denoising backbone $\Psi(\cdot)$ to extrapolate tri-planes.
The extrapolation is carried out in the latent tri-plane space. 
Formally, given a known block $P$ with latent code $z^{P} = \bigl\{z^{P}(i) | i\in\{1,2,3\}\bigr\}$ as a condition, 
and an empty block $Q$ that partially overlaps with $P$,
the goal is to generate the latent tri-plane 
$ z^{Q} = \bigl\{z^{Q}(i)| i\in\{1,2,3\} \bigr\}$ that can represent the new block.
For simplicity, this paper only considers the case where $Q$ is positioned by sliding along only one of the XYZ axes, which is sufficient for scene expansion.

\medskip
\noindent
\textbf{Plane-wise extrapolation.}
The tri-plane is a factored representation of a dense 3D volume.
The three planes are compressed but highly correlated,
which makes extrapolation on tri-planes a non-intuitive task.
To address this, 
as shown in Fig.~\ref{fig:pipe_triplane_extrapolation},
we factor tri-plane extrapolation into the extrapolation of three 2D planes separately,
and then utilize our 3D-aware denoising backbone, $\Psi$, to blend information from the three planes.
Specifically, given the $i$-th axis-align plane with $i\in\{1,2,3\}$, the overlap mask between plane $z^{P}(i)$ and plane $z^{Q}(i)$ is denoted as $O_i$.
Following Repaint~\cite{lugmayr2022repaint}, extrapolating $z^{P}(i)$ to obtain $z^{Q}(i)$ is realized by synchronizing the denoising process of $z^{Q}(i)$ using the noised version $z^{P}(i)$ inside the overlap mask $O_i$.  
Specifically, at step $t-1$, we obtain the noised $z^P_{t-1}(i)$ via 
\begin{equation}
z^P_{t-1}(i) \sim \mathcal{N}(\sqrt{\bar{\alpha}_t}z^P_0(i),(1-\bar{\alpha}_t)\textbf{I})    
\end{equation}
and the denoised  $z^Q_{t-1}(i)$  from previous step $t$ by 
\begin{equation}
z^Q_{t-1}(i) \sim  \mathcal{N}(\mu_{\psi}(z^Q_t(i),t),\Sigma_{\psi}(z^Q_t(i),t))  
\end{equation}
Then, $z^Q_{t-1}(i)$ is synchronized by
\begin{equation}
    z^Q_{t-1}(i) \gets \textnormal{Cat}\Bigl( z^P_{t-1}(i) \in O_i , \; z^Q_{t-1}(i) \notin O_i \Bigr) 
\end{equation}
where $\textnormal{Cat}(\cdot)$ refers to the tensor concatenation. 
However, as shown in Fig.~\ref{fig:pipe_triplane_extrapolation}, when $i=3$, the two planes $z^{P}(3)$ and $z^{Q}(3)$ are parallel to each other, and thus they do not have explicit overlap. 
We can only perform synchronization for planes $i\in\{1,2\}$.
Nevertheless, our denoising backbone $\Psi$ constructed using a sequence of self-attention layers is designed to identify cross-plane dependencies. 
This architecture allows the synchronized features in planes $\{1,2\}$ to be effectively propagated to the 3rd plane via attention layers throughout the denoising steps. 
We found in experiments that this approach successfully achieves meaningful 3D shape extrapolation.
The overall procedure for latent tri-plane extrapolation is outlined in Algorithm \ref{alg:cap}.

\begin{algorithm}[!ht]
\caption{Latent tri-plane extrapolation}\label{alg:cap}
\begin{algorithmic}
\State $ z^Q_T \sim \mathcal{N}(\textbf{0},\textbf{I}) $
\State $\textbf{for}\   t = T,...,1\  \textbf{do}$
\State\ \ \ \ \ \ \ \ $
z^P_{t-1} \sim \mathcal{N}(\sqrt{\bar{\alpha}_t}z^P_0,(1-\bar{\alpha}_t)\textbf{I}) 
$
\State\ \ \ \ \ \ \ \ $
z^Q_{t-1} \sim  \mathcal{N}(\mu_{\psi}(z^Q_t,t),\Sigma_{\psi}(z^Q_t,t))
$ 
\State\ \ \ \ \ \ \ \  $\textbf{for}\   i \in \{1,2\}\  \textbf{do}$
\State\ \ \ \ \ \ \ \ \ \ \ \  $z^Q_{t-1}(i) \gets \textnormal{Cat}\Bigl( z^P_{t-1}(i) \in O_i , z^Q_{t-1}(i) \notin O_i \Bigr)$
\State\ \ \ \ \ \ \ \  $\textbf{end for}$
\State$\textbf{end for}$
\State$return \ \ z^P_0, \ \ z^Q_0$
\end{algorithmic}
\end{algorithm}

\medskip
\noindent
\textbf{Resampling.}
We found that simply applying synchronization does not always yield semantically and geometrically consistent results. 
This is because the noise-adding process in overlapping regions does not take into account the newly generated parts of the tri-plane in the non-overlapping region, thereby introducing disharmony.
To address this issue, we leverage the resampling strategy as introduced in Repaint~\cite{lugmayr2022repaint}. 
Specifically, at certain steps of the denoising process, noise is added again to the output using the forward diffusion equation in~\ref{eqn:sample}, meaning the inference process is rolled back.
There are two hyper-parameters for resampling: 1) the roll-back step $J$, and 2) the number of resampling times $R$. 
In this paper, we set $J=100$ and conduct an ablation study on $R=\{0,1,2,3,7\}$.
Experimental results show that increasing the number of resampling times enhances the generation performance.

\begin{figure*}[!ht]
\includegraphics[width=0.95\textwidth]{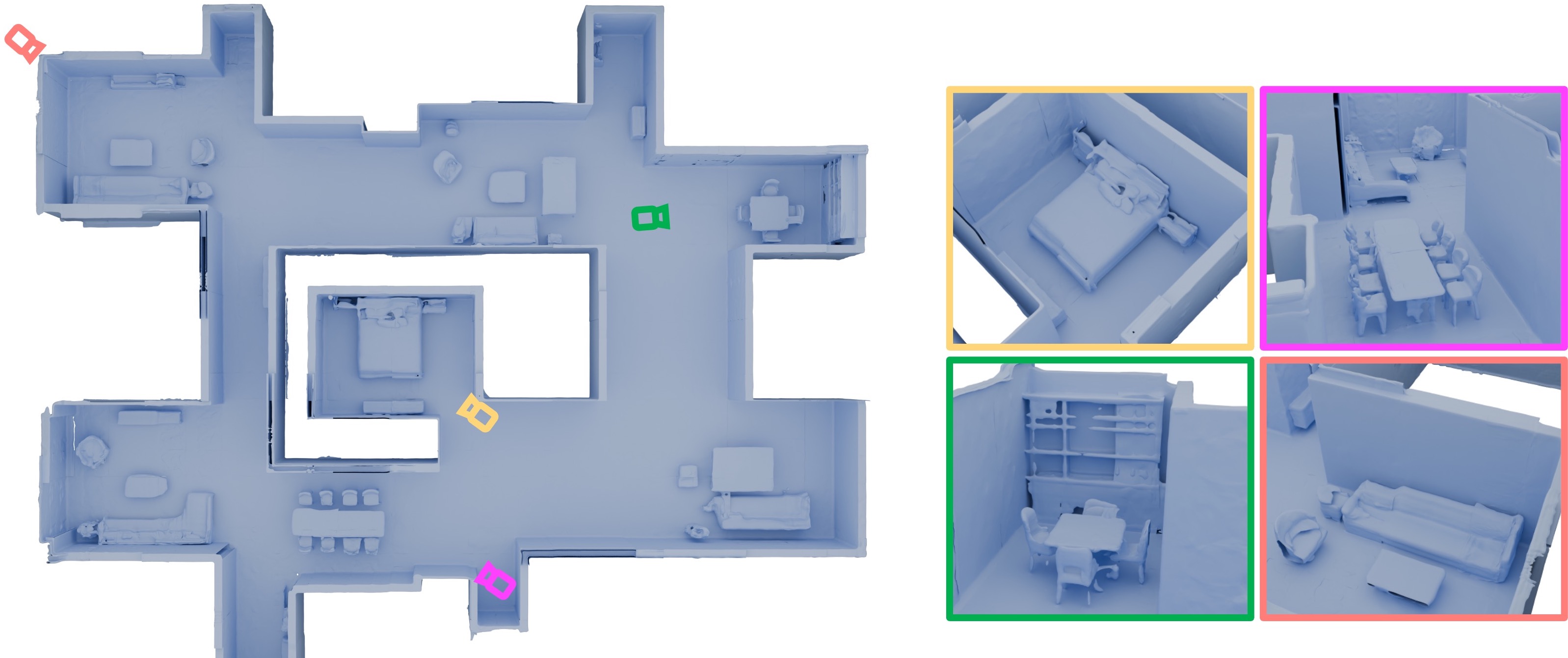} 
\caption{ Large room scene generation.}
\label{fig:room_large}
\end{figure*}
\begin{figure*}[!h]
\includegraphics[width=0.95\textwidth]{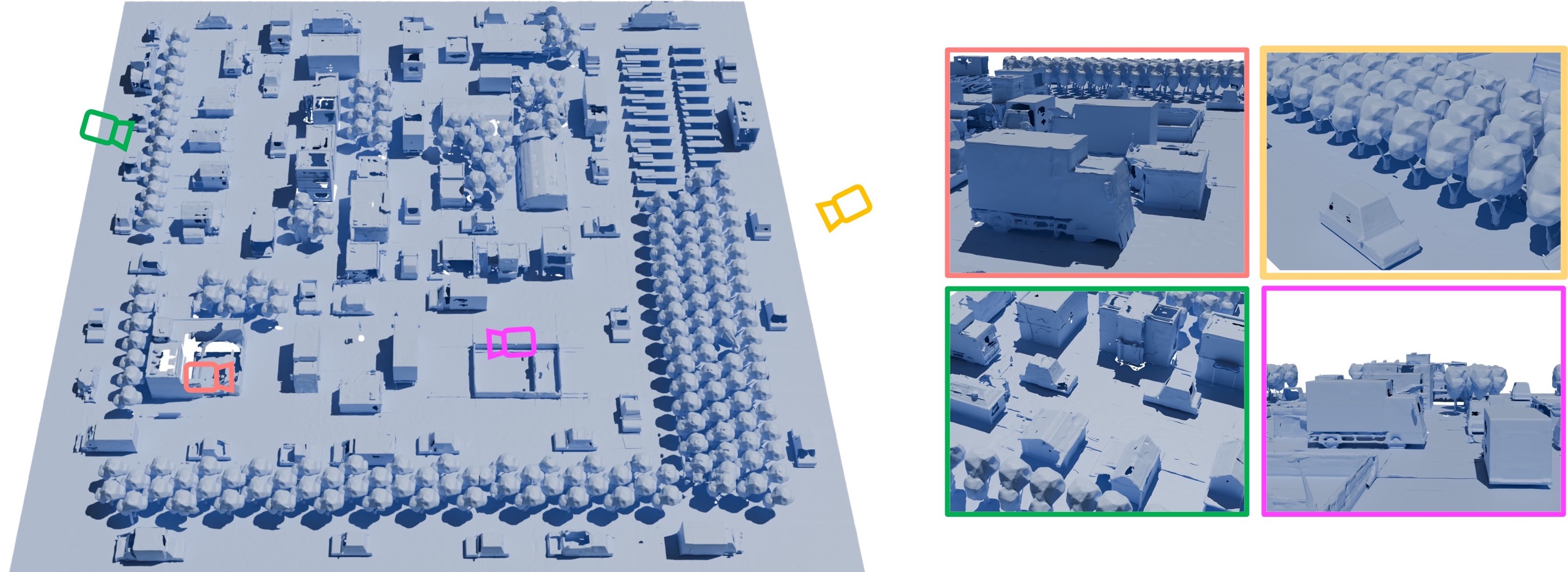} 
\caption{ Large city scene generation.}
\label{fig:city_large}
\end{figure*}

\subsection{Surface refinement with non-rigid registration} \label{sec:pp}
The synchronized tri-planes $z^{P}$ and $z^{Q}$ generate shapes that are semantically aligned; 
however, the latent space synchronization does not guarantee point-accurately aligned shapes, 
resulting in small visible seams.
To address this problem, we explicitly align the extracted surface mesh.
From the two latent codes $z^{P}$ and $z^{Q}$, we derive dense SDF volumes and run marching cubes to extract the surface meshes $S^{P}$ and $S^{Q}$.
We uniformly sample points on mesh triangles that lie inside the overlapping region, obtaining the point sets denoted as $\Omega^{ol}_P$ and $\Omega^{ol}_Q$.
Then, we uniformly sample points on triangles of $S^{Q}$ that lie outside of the overlapping region, resulting in a point set denoted as $\Omega^{new}_Q$.
We align $S^P$ and $S^Q$ by optimizing the non-rigid registration cost:
\begin{equation}
    \mathcal{L}_{nrr} = 
    \mathcal{L}_{CD} \Bigl( \mathcal{W} \bigl(\Omega^{ol}_P \bigr) , \Omega^{ol}_Q \Bigr)   
    +  
    \mathcal{L}_{CD} \Bigl( \mathcal{W} \bigl(\Omega^{new}_Q \bigr) , \Omega^{new}_Q \Bigr)
\end{equation}
where $\mathcal{L}_{CD}(\cdot)$ represents the Chamfer Distance between two point clouds, 
and $\mathcal{W}(\cdot)$ is the dense non-rigid warping function that predicts per-point transformations. 
$\mathcal{W}(\cdot)$ is based on NDP~\cite{li2022ndp}, which approximates scene deformation using hierarchical coarse-to-fine neural deformation fields. 
This non-rigid registration cost encourages the extrapolated mesh $S^Q$ to approximate the condition mesh $S^P$ as closely as possible within the overlapping region,
while maintaining its own structure in the non-overlapping region.

\subsection{Building unbounded large scenes with BlockFusion.}\label{sec:large_scene}
Based on Algorithm \ref{alg:cap}, one can construct large, unbounded scenes at any scale. The naive strategy for this purpose involves initially creating a block and then expanding the scene by extrapolating block by block in the sliding window fashion. 
However, this serial operations requires a significant amount of time.

Given that remote blocks are likely to be independent of each other, large scene generation can be executed in parallel. 
This process involves initially generating isolated seed blocks simultaneously, from which we extrapolate the remaining empty blocks, also in parallel.
Specifically, we first use sliding window to slice the world into small blocks, denoted as $\mathcal{B}$ = \{$B_{1},B_{2},...$\}, with overlaps between each pair of neighboring blocks  
We select a strided subset $\mathcal{B}^{seed}$ from those blocks.
We make sure blocks in $\mathcal{B}^{seed}$  should not overlap with each other. 
The complementary set of $\mathcal{B}^{seed}$ is denoted as $\mathcal{B}^{extra}$.
Blocks in $\mathcal{B}^{seed}$ are independently generated in parallel.
The rest empty blocks in $\mathcal{B}^{extra}$ are extrapolated from $\mathcal{B}^{seed}$.

\begin{figure*}[!t]
\centering
\includegraphics[width=1\textwidth]{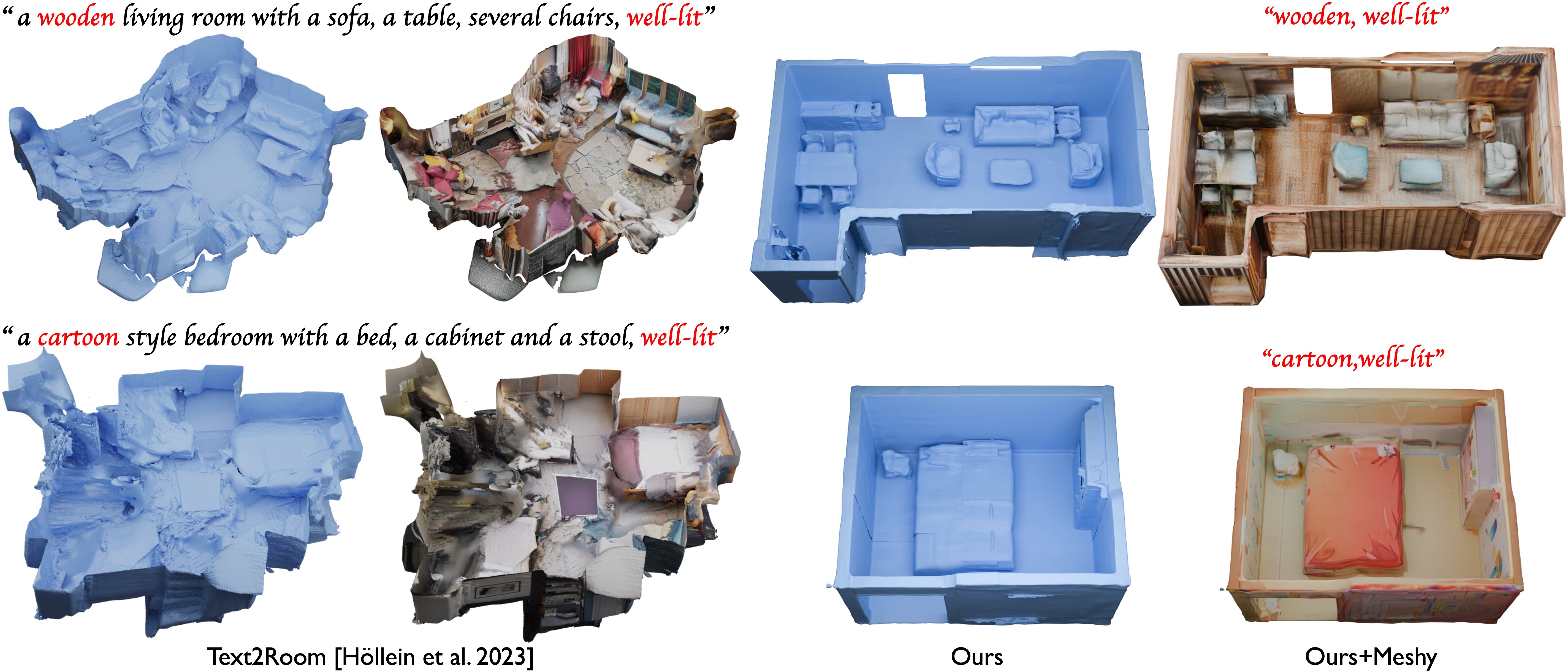}
\caption{
\textbf{Qualitative room generation results.}
Text2Room~\cite{hollein2023text2room} generates distorted shapes and cannot accurately respond to the number of objects in the scene. For instance, when given the prompt "one bed", it generates multiple beds.
In contrast, BlockFusion produces higher-quality shapes and correctly responds to numerical prompts.
}

\label{fig:compare_text2room}
\end{figure*}

\section{Experimental Results}

\subsection{Implementation details.}

\medskip
\noindent
\textbf{Water-tight remeshing.}
We use 3D scene meshes for network training.
These scene meshes are typically created by 3D artists and are not always guaranteed to be watertight. 
We transform the raw meshes into watertight ones using Blender's \textit{voxel remeshing} tool.
After remeshing, the object has a clearly defined inside and outside, 
which is essential for training a continuous neural field representation.

\medskip
\noindent
\textbf{Datasets.} We test our algorithm on three different types of scenes: room, city, and village.
Room scene data is obtained from 3DFront~\cite{3dfront} and 3D-FUTURE~\cite{fu20213d}, which contains 18,968 indoor scenes with 34 classes of indoor objects. 
We obtain 57K random crops from 3DFront, with each block size set to $3.2^3$ cubic meters. 
We filtered out empty rooms and rooms with less than 5 objects and finally got 9123 rooms.
For simplicity, we regroup the objects in 3D-FUTURE based on their similarities into 9 classes: "floor", "wall", "chair", "cabinet", "sofa", "table", "lighting", "bed", and "stool".
The city and village scenes are designed by artists, 
from each, we obtain 10K blocks. 
The block sizes are set to $12^3$ and $15^3$ cubic meters, respectively.
The layout labels for the village are "pine", "cypress tree", "ground", and "houses", and for the city, they are "road", "tree", "solar panels", "cars", "houses".
Note that all the blocks are cropped at random, 
and the testing blocks are never exposed to the model during training. 
During inference, the input semantic layout is created using an easy-to-use GUI, where user can place bounding boxes to indicate objects such as "car" and "tree", or draw lines/contours to indicate continuous areas such as "wall" and "road". \\

Our method is implemented using Pytorch and trained on Nvidia V100 GPU.
For the 3D Front dataset with 57K cropped blocks, raw tri-plane fitting, auto-encoder training, and diffusion training take 4750, 768, and 384 GPU hours, respectively.
VAE and diffusion training require 8 GPUs. Tri-plane fitting and BlockFusion inference can run on one GPU. 
Running a single tri-plane extrapolation under layout conditions costs 6 minutes.
With the large scene generation strategy described in Sec.~\ref{sec:large_scene}, producing the large indoor scene in Fig.~\ref{fig:room_large} takes around 3 hours.

\subsection{Evaluation Metrics}

\medskip
\noindent
\textbf{Reconstruction metric.}
We evaluate the reconstruction quality using the Chamfer Distance (CD) at $10^{-3}$ scale, Surface Normal Error ($E_{NRM}$) in degrees, and Surface SDF error ($E_{SDF}$) in centimeters.
    
\medskip
\noindent
\textbf{Unconditioned generation metric.}
The evaluation of unconditional 3D shape synthesis presents inherent challenges due to the lack of direct ground truth correspondence. Therefore, we resort to well-established metrics for evaluation, in line with previous works~\cite{diffusion-sdf, zeng2022lion,siddiqui2023meshgpt}. These metrics include Minimum Matching Distance (MMD), Coverage (COV), and 1-Nearest-Neighbor Accuracy (1-NNA). For MMD, lower is better; for COV, higher is better; for 1-NNA, 50\% is the optimal. We employ the Chamfer Distance (CD) and EMD (Earth Mover's Distance) as the distance measure for computing these metrics. 
More comprehensive details about these metrics are available in the respective literature.

\medskip
\noindent
\textbf{User study metric.}
We carried out a user study involving 48 participants who were asked to rate the scene generation results based on Perceptual Quality (PQ) and Structure Completeness (SC) of the entire scene, using a scale from 1 to 5.
This is done in two modes: textured mode (T-) and geometry-only mode (G-). 
In the Textured Mode, participants viewed a textured mesh, while in the Geometry-Only Mode, the texture was replaced with a monochrome material to emphasize the geometry.
As a result, we derived four metrics from this study: T-PQ, T-SC, G-PQ and G-SC.
The results are presented in Table
\ref{tab:indoor_eval_geo} and~\ref{tab:indoor_eval_tex}.

\begin{table}[!ht]
\caption{Quantitative indoor scene generation results of pure shape.}
\resizebox{0.75\columnwidth}{!}{
\begin{tabular}{lcc}
\toprule
& GPQ$\uparrow$                 & GSC$\uparrow$               \\
\midrule          
Text2Room~\cite{hollein2023text2room} &                    1.92&                  1.92\\
\midrule
Ours      &                    \textbf{4.44}&                  \textbf{4.58}\\
\bottomrule
\end{tabular}
}
\label{tab:indoor_eval_geo}

\end{table}
\begin{table}[!ht]
\caption{Quantitative indoor scene generation results of textured shape.}
\resizebox{0.75\columnwidth}{!}{
\begin{tabular}{lcc}
\toprule
& TPQ$\uparrow$              & TSC$\uparrow$              \\
\midrule          
Text2Room~\cite{hollein2023text2room} &                 2.27&                 2.29\\
\midrule
Ours + Meshy&                 \textbf{4.08}&                 \textbf{4.17}\\
\bottomrule
\end{tabular}
}
\label{tab:indoor_eval_tex}
\end{table}

\begin{figure*}[!t]
\newcolumntype{C}[1]{>{\centering\let\newline\\\arraybackslash\hspace{0pt}}m{#1}}
\newcommand{\cw}{85pt}
\centering
\includegraphics[width=1\textwidth]{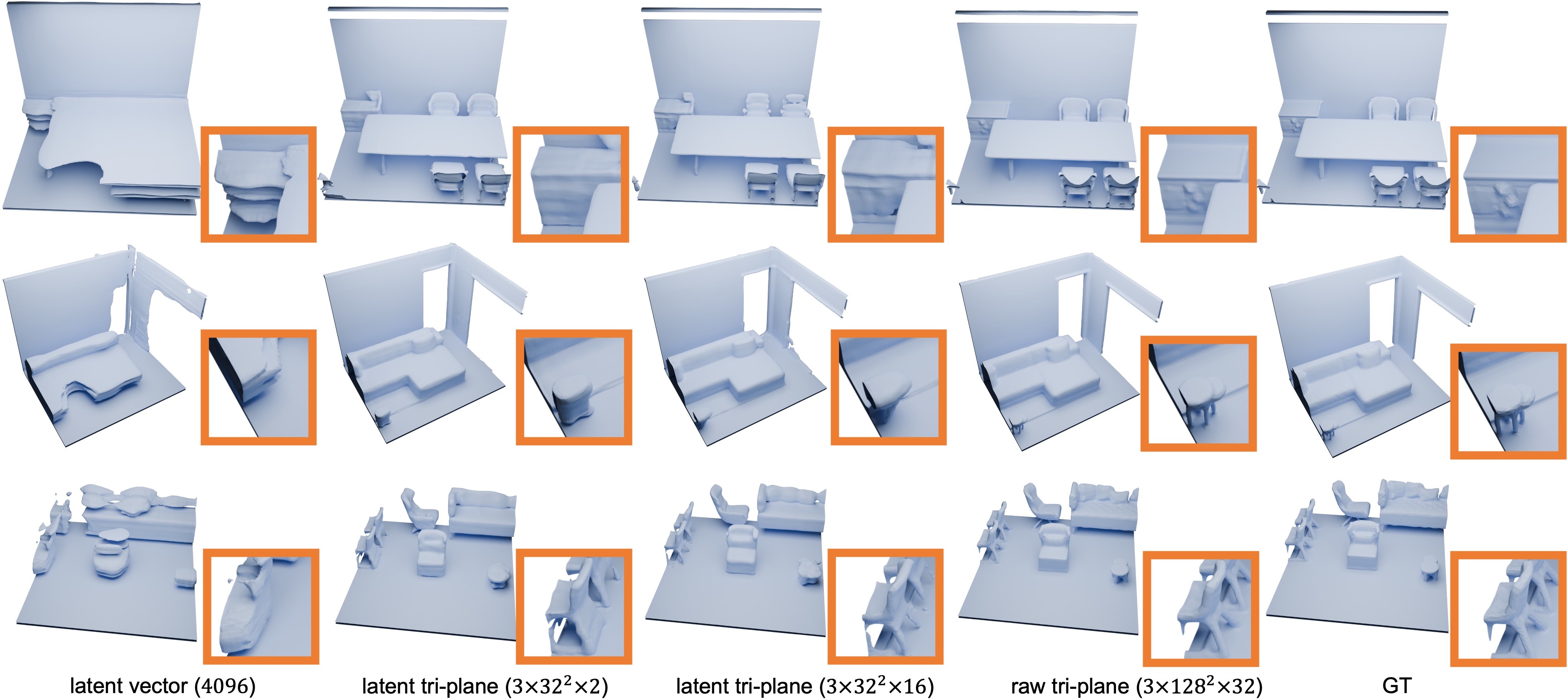} 
\caption{ \textbf{Qualitative block reconstruction results.}
The raw tri-plane can faithfully represent the ground truth (GT) mesh without any issues.
The latent tri-planes with 2 channels significantly reduce the total number of parameters, while only causing moderate shape degradation.
The latent vector struggles to represent 3D scenes accurately.
}
\label{fig:block_recon}
\end{figure*}
 
\subsection{Comparison with SOTA.}

\medskip
\noindent
\textbf{Single block generation.}
We regard NFD~\cite{nfd} as the baseline for single block generation. 
NFD is also based on tri-plane diffusion. 
However, they use occupancy values to represent shapes, whereas we employ SDF. 
We retrain NFD on our indoor scene blocks before evaluation.
Tab.~\ref{tab:uncond_gen} and Fig.~\ref{fig:uncond_gen} show the results of unconditioned indoor block generation.
Quantitatively, our method ($z_0$ target) significantly outperforms NFD, with a $29.17\%$ and $23.67\%$ increase in coverage (Cov) scores under the CD and EMD metrics, respectively. 
Qualitatively, NFD is unable to generate meaningful shapes.

\begin{table}[!h]
\caption{Quantitative unconditional generation results for indoor blocks.}
\resizebox{1\columnwidth}{!}{
\begin{tabular}{lcccccc}
\toprule
& \multicolumn{2}{c}{MMD $\downarrow$} & \multicolumn{2}{c}{COV(\%,$\uparrow$)} & \multicolumn{2}{c}{1-NN(\%,$\downarrow$)}\\
\cmidrule{2-7}
& CD             & EMD              & CD                & EMD            &
CD               & EMD\\
\midrule          
NFD~\cite{nfd}   &     0.0445            &     0.2363            &    22.66                & 29.66 & 89.08 & 83.25                \\
\midrule
 Raw tri-plane Diff. w/ $z_0$ (Ours)       &   0.0544              &      0.2744           &     23.99               &       27.50
 & 89.91 & 88.50\\
\midrule     
 Latent tri-plane Diff. w/ $z_0$ (Ours)       &  \textbf{0.0324}               &     0.1884           &          \textbf{51.83}          &   53.33    & 70.66 & 60.08           \\
\midrule 
  Latent tri-plane Diff. w/ $noise$ (Ours)       &  0.0326               &     \textbf{0.1865}            &          50.49          &   \textbf{55.66}    & \textbf{69.74} & \textbf{56.66}           \\
\bottomrule
\end{tabular}
}
\label{tab:uncond_gen}
\end{table}

\medskip
\noindent
\textbf{Indoor scene generation.}
We consider Text2Room~\cite{hollein2023text2room} as the baseline for indoor scene generation. 
Text2Room takes text prompt as input whereas ours is based on 2D layout map.
For a fair comparison, we describe our input room layout using natural language and then concatenate it as part of the text prompt for Text2Room. 
Since our method does not directly generate textured meshes, we leverage an off-the-shelf text-to-texture generation tool, Meshy~\footnote{https://www.meshy.ai/, to produce textures for our mesh.
}
Meshy utilizes the same text prompt as Text2Room.
To enhance the texture generation results of Meshy, we combine all blocks into a single entity using Blender's voxel remeshing tool.
Tab.~\ref{tab:indoor_eval_geo}, Tab.~\ref{tab:indoor_eval_tex} and Fig.~\ref{fig:compare_text2room} show the results of room generation. 
Qualitatively, due to the use of monocular depth estimation, the shape of Text2Room appears distorted, 
while BlockFusion produces significantly better room shapes. 
In addition, Text2Room cannot precisely react to the text prompt; it generates duplicate beds, while the prompt is "one bed". 
By leveraging layout control, our method can precisely determine the number of beds in the room. 
By leveraging Meshy, our approach also produces textures comparable to Text2Room.
Quantitatively, under a five-point system, Blockfusion are leading by $2.52$ points in geometric perceptual quality and $2.66$ points in geometric structure completeness respectively. 
In the case of textured generation, Blockfusion are leading by $1.81$ points in textured perceptual quality and $1.88$ points in textured structure completeness respectively. 

\begin{figure*}[!h]
\centering
\includegraphics[width=1\textwidth]{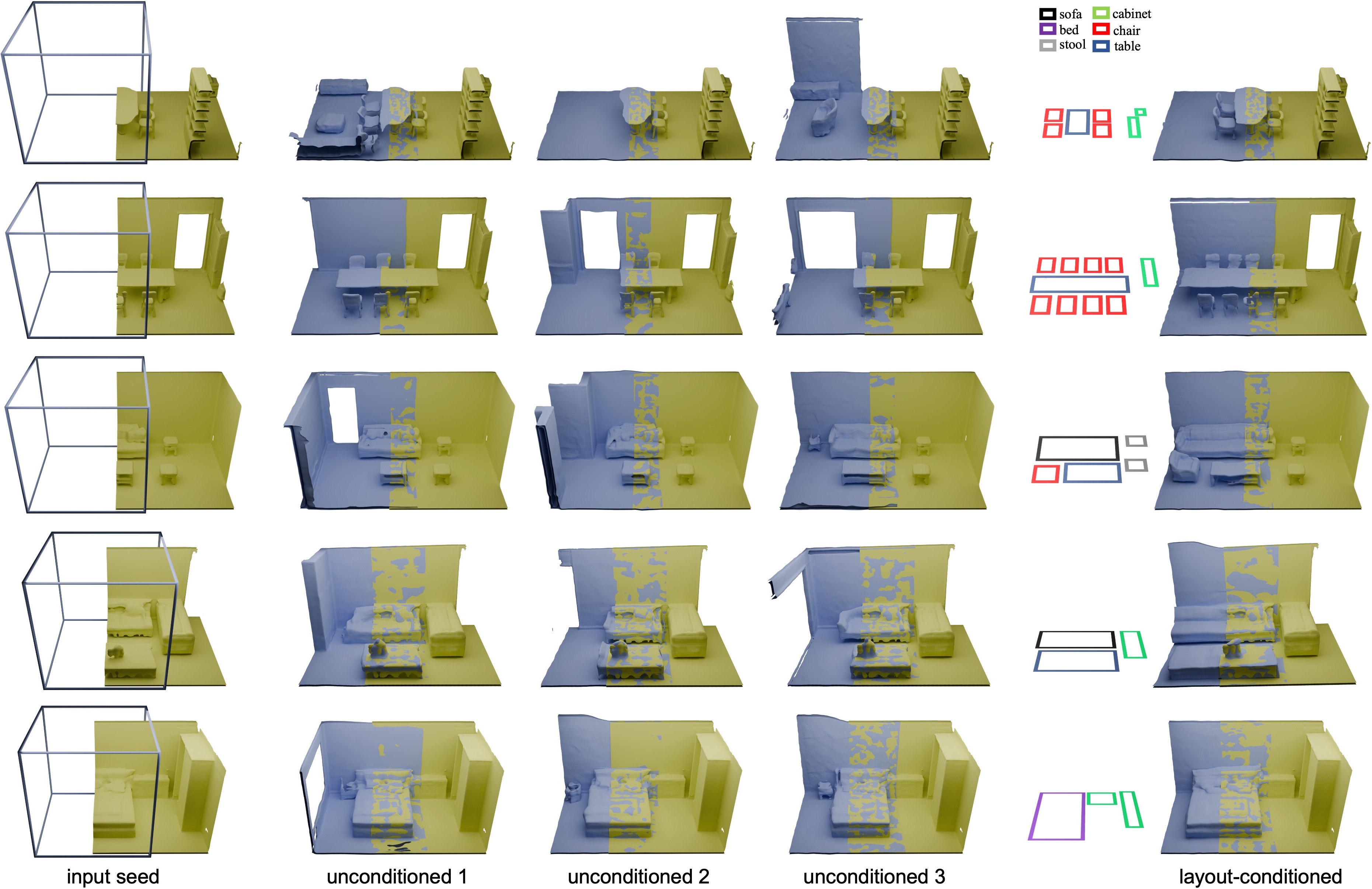} 
\caption{ 
\textbf{Qualitative results of tri-plane extrapolation.}
The 3D box shows the block to extrapolate. 
The overlap ratios are $25\%$ for top three rows and $50\%$ for bottom three rows.
}
\label{fig:extra}
\end{figure*}

\subsection{Ablation study}

\medskip
\noindent
\textbf{Shape reconstruction quality: latent tri-plane vs raw-tri-plane.}
Figure~\ref{fig:block_recon} and Table~\ref{tab:block_recon} display the qualitative and quantitative room block reconstruction results using different representations. 
The raw tri-plane can accurately represent the ground truth (GT) mesh without any issues. 
Compared to the raw tri-plane, the latent tri-planes with 2 channels manage to reduce $99.6\%$ of the data bits while still maintaining decent shape representation power. 
Using a similar data compression rate, the 4096-dimensional latent vector cannot produce any reasonable shape. 
Considerably, the raw tri-plane is a redundant 3D representation.
However, when we attempted to use fewer feature channels and resolution during the raw tri-plane fitting, we observed a considerable decline in the quality of shape reconstruction, as depicted in the first two rows of Table~\ref{tab:block_recon}.
This observation demonstrates the necessity of using an auto-encoder for tri-plane compression.
\begin{table}[!h]
\caption{Quantitative reconstruction results for indoor blocks.
CPR: compression rate w.r.t the raw tri-plane at resolution $3\times128^2\times32$ .
The units are CD with scale $10^{-3}$ , $E_{NRM}$ in degrees, and $E_{SDF}$ in centimeters.
}
\resizebox{1\columnwidth}{!}{
\begin{tabular}{llcccc}
\toprule
 &Dimension & CPR  & E$_{NRM}\downarrow$  & E$_{SDF}\downarrow$ & CD$\downarrow$ \\
\midrule
\multirow{2}{*}{Raw tri-plane}& $3\times128^2\times32$    & 0\%  & \textbf{5.39}   &  \textbf{0.0223}        &  \textbf{3.952}  \\
\cmidrule{2-6} &$3\times32^2\times2$ & 99.60\%   & 30.15   &  0.4290        &  5.560 \\
\midrule
\multirow{2}{*}{Latent tri-plane} & $3\times32^2\times16$ &  96.87\%    &    14.64          &    0.2231      & 4.070    \\
\cmidrule{2-6}
& $3\times32^2\times2$ &  99.60\%   &    15.61         &   0.2523        &  4.097   \\
\midrule
Latent vector& 4096&  99.73\%   &   18.77     &    0.3395   & 5.431\\
\bottomrule
\end{tabular}
}
\label{tab:block_recon}
\end{table}

\medskip
\noindent
\textbf{Shape generation quality: latent tri-plane vs raw-tri-plane.}
Figure~\ref{fig:uncond_gen} and Table~\ref{tab:uncond_gen} display the qualitative and quantitative unconditional room block generation results.
The latent tri-plane diffusion ($z_0$) shows significantly better results, with a $27.84\%$ and $25.83\%$ increase in Coverage scores under the CD and EMD metrics respectively.
Qualitatively, raw tri-plane diffusion can not produce any reasonable results. 
In conclusion, compared to the raw tri-plane, the latent tri-plane retains decent shape representation capacity while serving as a superior proxy for shape generation.

\medskip
\noindent
\textbf{How does the layout condition impact the generation process?}
As illustrated in Fig.~\ref{fig:extra}, unconditioned generation can produce multiple extrapolation results, while the conditioned version generally converges to the layout guidance.
Nonetheless, we found that layout conditions can dictate the overall placement of objects but not the intricate details of their shapes. 
This implies that various shapes can be achieved under the same layout conditions.
An example of this phenomenon can be observed in the supplementary video, where we demonstrate the generation of different sofas while still adhering to the specified layout conditions.
This showcases the flexibility and adaptability of our approach in generating diverse and unique scene elements while maintaining consistency with the given layout constraints.

\begin{figure}[!h]
\centering
\includegraphics[width=0.48\textwidth]{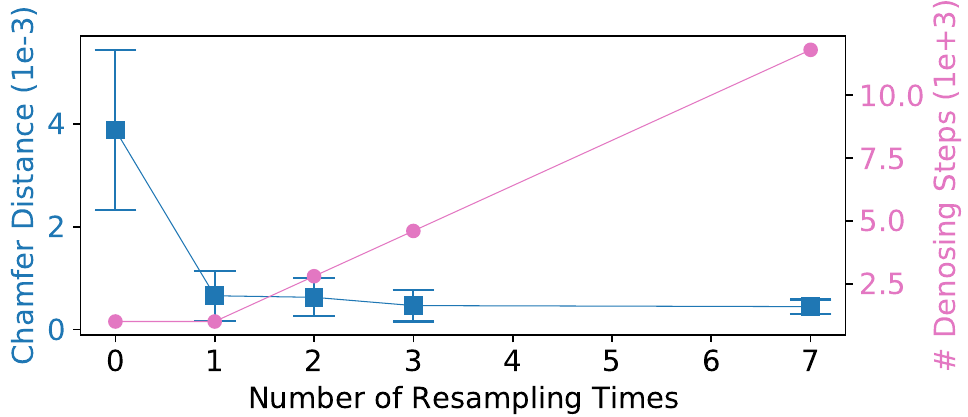} 
\caption{
\textbf{Layout-conditioned tri-plane extrapolation with different resampling times ($R$).}
The Chamfer Distance is calculated based on point sets sampled from the two block meshes within their overlapping region.
The shape consistency significantly improves after 1-time synchronization, and employing additional synchronization steps (i.e. resampling) further enhances shape consistency.
$R=0$ means no synchronizations.
}
\label{fig:Resample}
\end{figure}

\medskip
\noindent
\textbf{How does resampling affect the extrapolation?}
Fig.~\ref{fig:Resample} shows the shape synchronization results of layout-conditioned tri-plane extrapolation.
We tested different resampling times with $R=\{1,2,3,7\}$.
The Chamfer Distance drops steadily with more resampling steps and stabilizes after 3 resamplings,
where the variance in the Chamfer Distance also converges.
This suggests that augmenting the number of resampling times can improve the quality of synchronization results.
For clarity, $R=0$ means that we do not perform synchronizations, i,e. the two blocks are generated independently while adhering to the shared layout conditions. 
Note that in this case, the Chamfer Distance is extremely high, indicating that using layout conditioning alone does not ensure consistent geometry between blocks.

\begin{figure}[!t]
\centering
\includegraphics[width=0.49\textwidth]{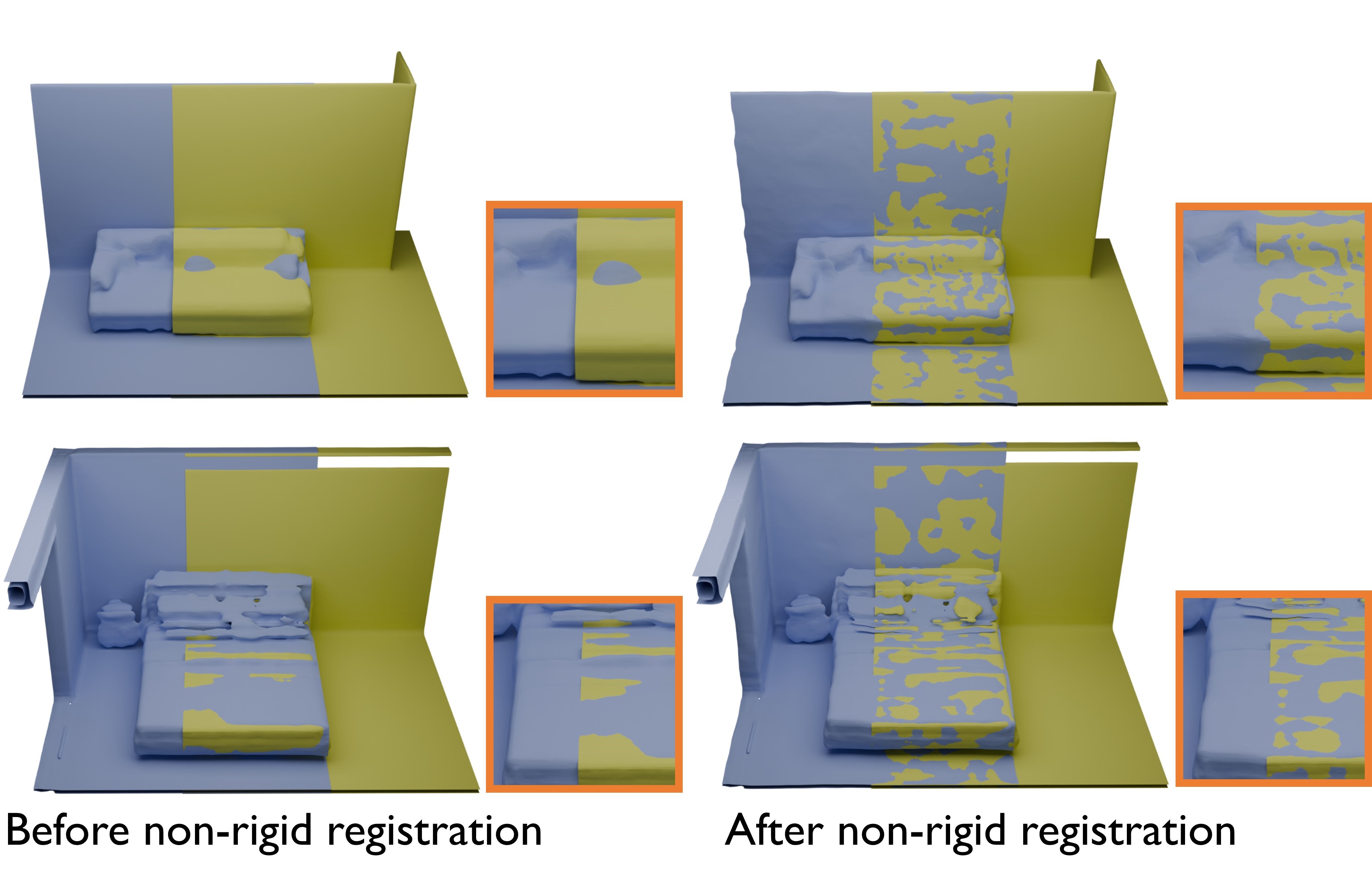} 
\caption{ Left: latent tri-plane extrapolation result (the seams are more visible when zoomed in), rights: after applying non-rigid registration. }
\label{fig:abl_postprocessing}
\end{figure}

\medskip
\noindent
\textbf{Is non-rigid registration-based post-processing necessary?}
Yes.
Latent tri-plane extrapolation generates semantically and geometrically reasonable transitions.
However, since the high-frequency, imperceptible details are abstracted away by the auto-encoder,
extrapolation in the latent tri-plane space inevitably results in minor seams.
As shown in Fig.~\ref{fig:abl_postprocessing}, non-rigid registration-based post-processing can effectively mitigate this issue.

\begin{figure}[!t]
\centering
\includegraphics[width=0.46\textwidth]{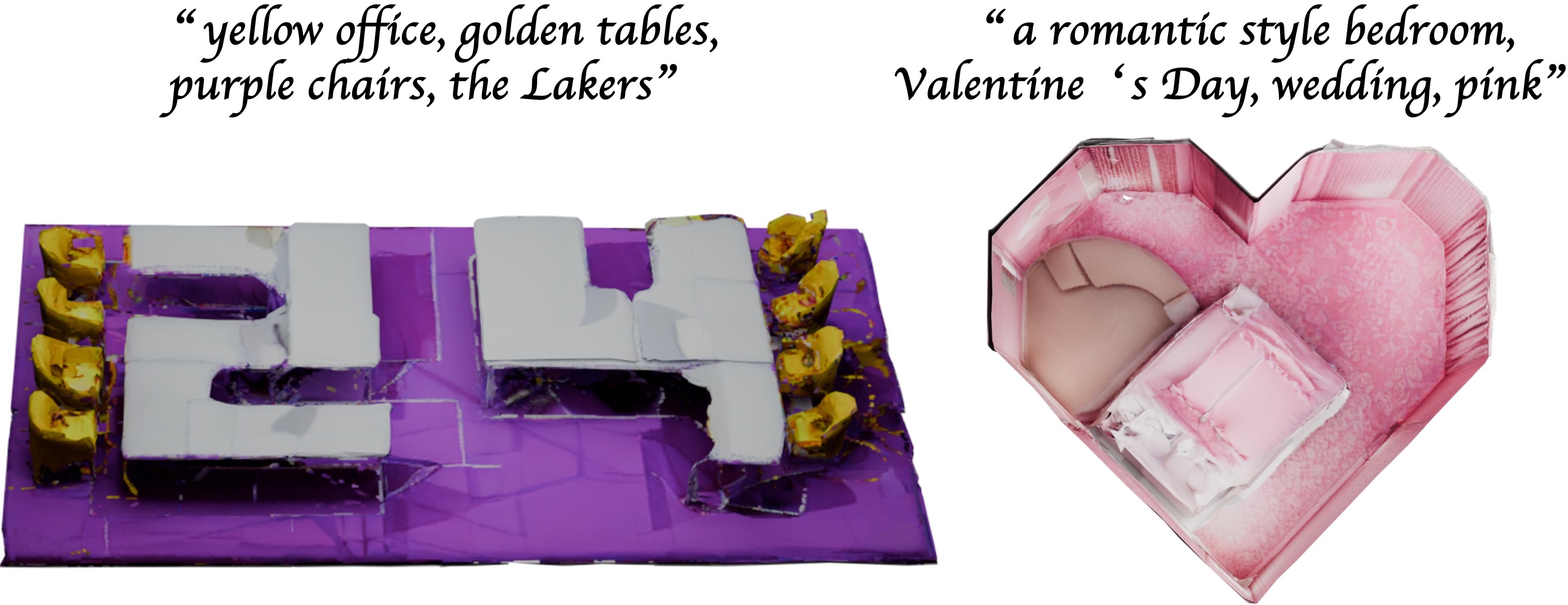} 
\caption{ \textbf{Using layout control to create rooms that do not exist in the training set.}
The textures are generated using Text2tex~\cite{chen2023text2tex} using the corresponding text prompt.
}
\label{fig:contentcreation}
\end{figure}

\medskip
\noindent
\textbf{Does BlockFusion posses creativity?}
BlockFusion does generate novel shapes that do not exist in the training dataset. 
This primarily arises from its ability to rearrange existing elements in novel ways. 
For instance, as shown in Fig.~\ref{fig:contentcreation}, BlockFusion manages to generate a new table shaped like the number "24" and a novel room shaped like a heart. 
This is made possible by its ability to re-combine basic shapes, such as fractions of tables and walls, under layout guidance.
This demonstrates the potential of BlockFusion as a powerful tool for generating diverse and visually appealing scenes.

\medskip
\noindent
\textbf{Predicting $z_0$ vs Predicting noise.}
In Section~\ref{sec:diffusion}, we adapt the strategy that predicts $z_0$ with $\Psi$ during the reverse diffusion process.
This is in contrast to the vanilla DDPM~\cite{ho2020denoising} which predict the noise. 
We conduct a quantitative ablation study to compare the two strategies. 
As shown in Table~\ref{tab:uncond_gen}, predicting $z_0$ and predicting noise as targets achieves comparable results in terms of unconditional generation.

\subsection{Large Scene Generation.}
We showcase the capability of BlockFusion for large scene generation. 
The results are displayed in Fig.~\ref{fig:teaser},~\ref{fig:room_large}, and~\ref{fig:city_large}, for village, city, and room scenes, respectively. 
The generation process is conditioned on layout maps that are created using an easy-to-use graphical user interface (GUI). 
It is important to emphasize that the scope of the scenes can be expanded  infinitely.
We believe that BlockFusion is the first method capable of generating 3D scenes at such a large scale while maintaining a high level of shape quality.

\section{Conclusion and Discussion}

Experiments show the proposed BlockFusion is capable of generating diverse, geometrically consistent, and unbounded large 3D scenes with high-quality geometry in both indoor and outdoor scenarios.
The generated mesh can be seamlessly integrated with off-the-shelf texture generation tools, yielding textured results with visually pleasing appearance.
We believe this approach represents an important step towards fully automated, industry-quality, large-scale 3D content generation. 

The expansive nature of BlockFusion allows it to serve as a map generator for open-world games. 
We integrate BlockFusion to Unity to develop such an open-world game, where players can roam and explore the world freely without being restricted by a predetermined world boundary. 
A demo of this can be found in the supplementary video.

\paragraph{\textbf{Advantage over procedural generation.}} Procedural Content Generation (PCG) is a complicated system that heavily depends on expertise. It requires carefully crafted rules to generate meaningful 3D scene. These rules must be re-programmed when transitioning between different scene styles. In contrast, BlockFusion is rule-programming-free and easy to use. It learns the distribution of scenes directly from data, and the generation process is fully automated. 

\paragraph{\textbf{Limitations.}}  The current implementation of BlockFusion faces several limitations. 
Our method may fail to generate very fine geometric details in the scene, such as the legs of a chair.
This issue primarily stems from the limited resolution used for the tri-planes. A possible solution is to adopt tri-plane super-resolution. 
Moreover, the bounding box condition can only control the approximate placement of objects, not their orientations. 
We believe that precise orientation control could be achieved by training diffusion conditioning on both the bounding box map and an object orientation map. 
This orientation map can also be easily obtained from user instructions.
Lastly, while we have demonstrated textured mesh results on small scenes, 
the task of generating globally consistent textures for large scene meshes is both a challenging and intriguing future endeavor.

\bibliographystyle{ACM-Reference-Format}
\bibliography{main}

\clearpage

\begin{appendix}

\begin{samepage}

\section{Qualitative room generation}

To reduce variables in the experiment, we validated the coloring of generated geometry by Meshy for Text2room~\cite{hollein2023text2room}. Additional qualitative and quantitative results are presented in Fig. \ref{fig:supqual} and Table \ref{tab:sup_eval_tex}. A total of 9 people are included in this user study.

\begin{figure}[!h]
\centering
\includegraphics[width=0.49\textwidth]{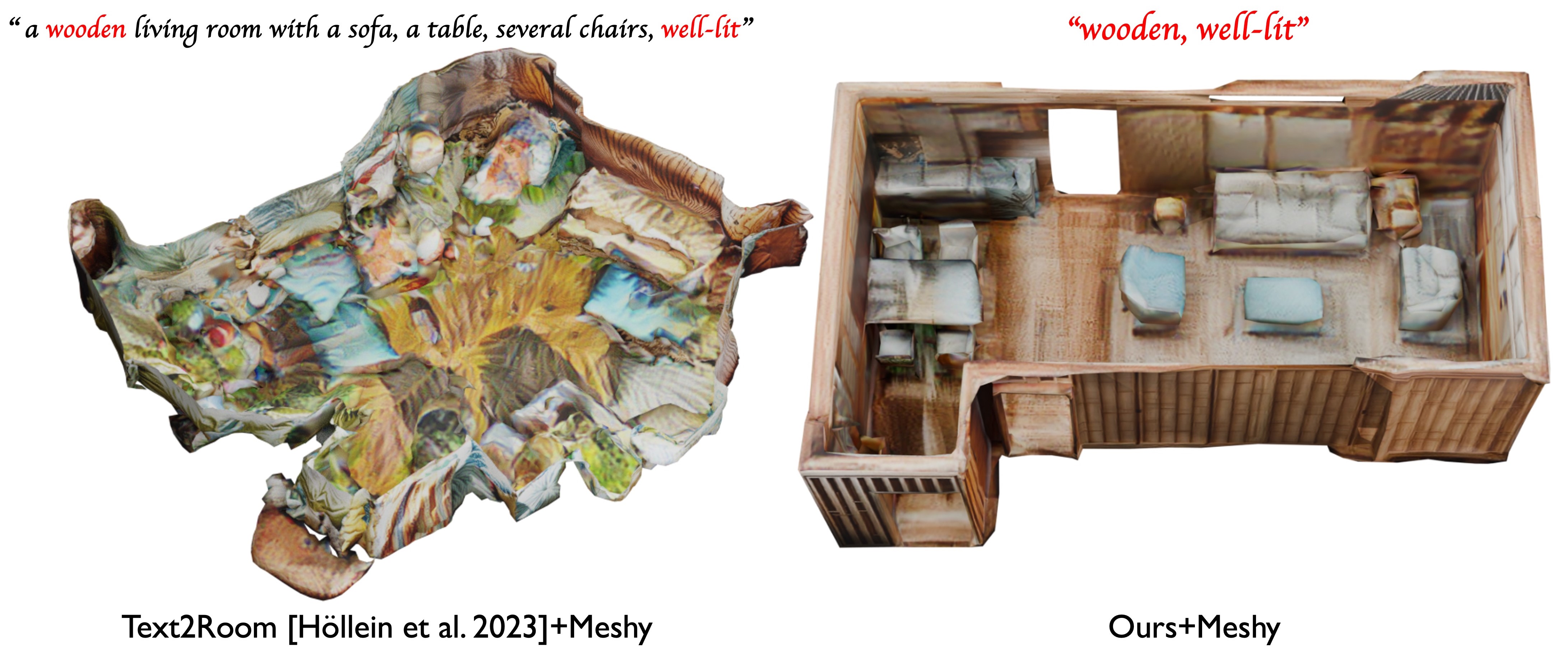}
\caption{
\textbf{Supplementary room generation results.}
Text2Room~\cite{hollein2023text2room} generates distorted shapes. While using Meshy to generate texture, it
cannot correspond well to the input prompts.
}

\label{fig:supqual}
\end{figure}

\begin{table}[h]
\caption{Quantitative indoor scene generation results of textured shape.}
\resizebox{0.7\columnwidth}{!}{
\begin{tabular}{lcc}
\toprule
& TPQ$\uparrow$              & TSC$\uparrow$              \\
\midrule          
Text2Room + Meshy &                 1.22&                 1.22\\
\midrule
Ours + Meshy&                 \textbf{4.56}&                 \textbf{4.67}\\
\bottomrule
\end{tabular}
}
\label{tab:sup_eval_tex}
\end{table}

\section{VAE Structure}

Fig. \ref{fig:vaeenc} and \ref{fig:vaedec} details our VAE for compressing raw tri-plane to latent tri-plane space as described in Section 3.3. 
Parameters of ResNet \cite{he2016deep} blocks
are given as (\#in channels, \#out channels). Parameters of Transformer \cite{dosovitskiy2020image} layers are given as (\# channels)
All shapes are denoted as (\#batch,\#channel,Height,Width). 
$C$  refers to the number of channels compressed into the latent tri-plane space.
Feature Pyramid Networks \cite{lin2017feature} are hired to aggregate multi-scale features.
The tri-planes are unfolded into
three independent planes to run convolutions separately. 
Transformer layers are leveraged to achieve cross-plane dependence.

\begin{figure}[h]
\centering
\includegraphics[width=.4\textwidth]{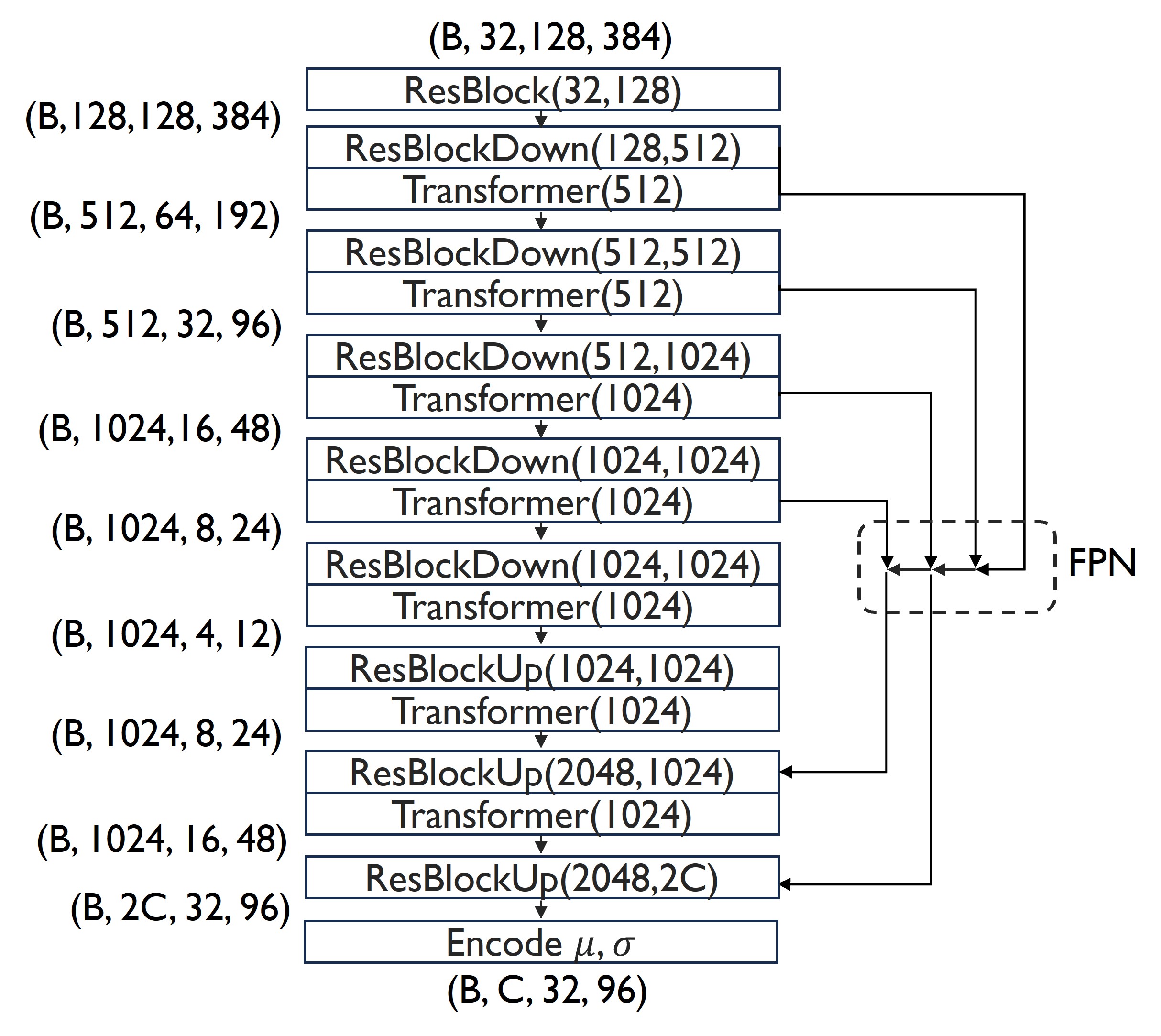} 
\caption{ 
\textbf{VAE structure.}
VAE enocder structure.
}
\label{fig:vaeenc}
\end{figure}

\begin{figure}[h]
\centering
\includegraphics[width=.4\textwidth]{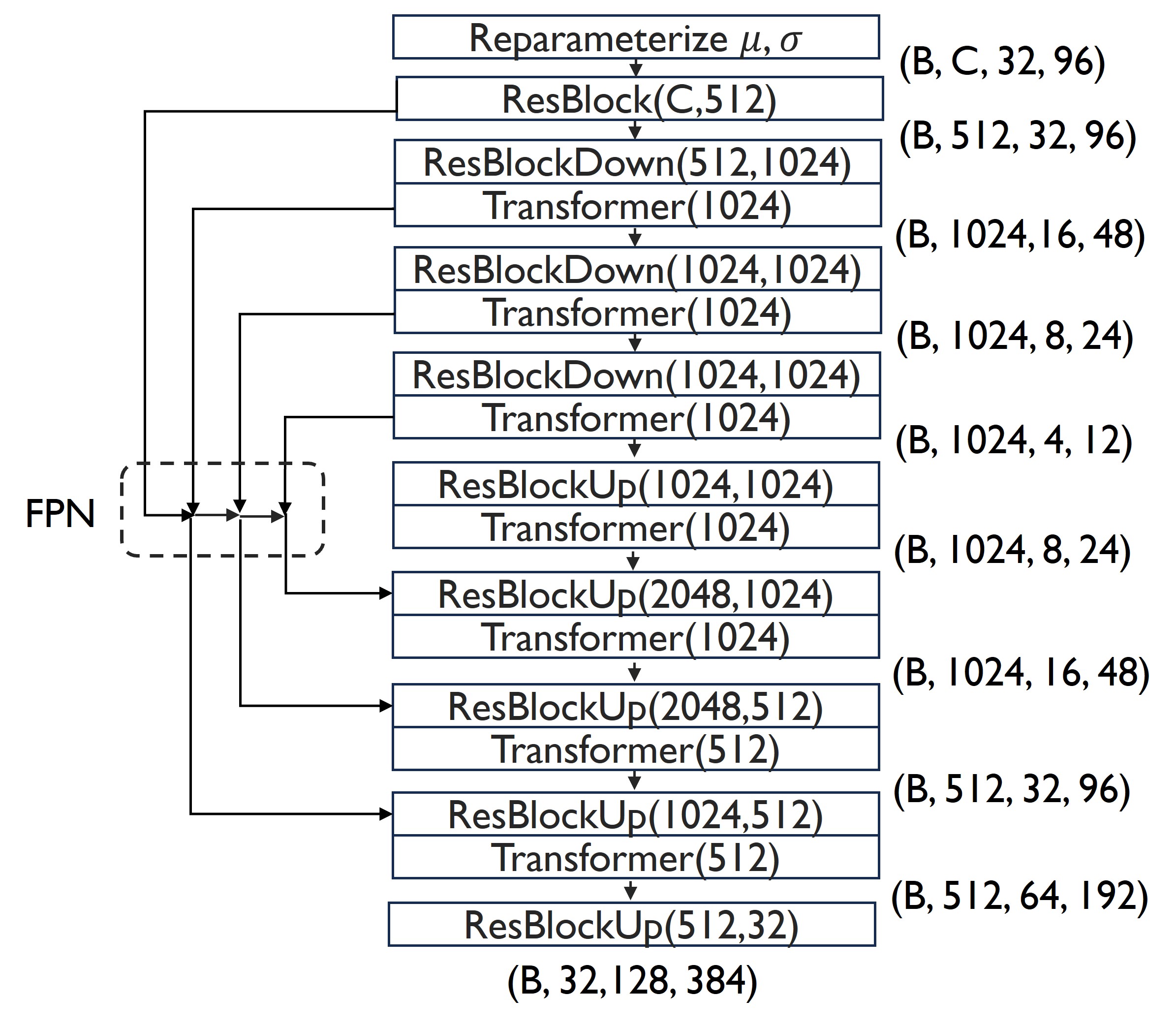} 
\caption{ 
\textbf{VAE structure.}
VAE decoder structure.
}
\label{fig:vaedec}
\end{figure}

\end{samepage}
\end{appendix}

\end{document}